\documentclass[10pt,twocolumn,letterpaper]{article}

\usepackage{cvpr}              
\usepackage{algorithm}
\usepackage{tabularx}
\usepackage{algpseudocode}
\usepackage[accsupp]{axessibility}

\usepackage{multirow}
\usepackage{placeins}
\usepackage{makecell}
\usepackage[table]{xcolor}

\usepackage{placeins}

\usepackage{blindtext}
\usepackage{xcolor}
\usepackage{bbm}
\usepackage{letltxmacro,xparse}
\LetLtxMacro{\blindtextblindtext}{\blindtext}
\LetLtxMacro{\blindtextBlindtext}{\Blindtext}
\RenewDocumentCommand{\blindtext}{O{\value{blindtext}}}{%
  \begingroup\color{gray}\blindtextblindtext[#1]\endgroup
}
\RenewDocumentCommand{\Blindtext}{O{\value{blindtext}}O{\value{Blindtext}}}{%
  \begingroup\color{gray}\blindtextBlindtext[#1][#2]\endgroup
}

\usepackage{xspace}  
\newcommand{\pretraining}{pre-training\xspace}
\newcommand{\finetuning}{fine-tuning\xspace}

\usepackage{pifont}
\newcommand{\cmark}{\ding{51}} 
\newcommand{\xmark}{\ding{55}} 

\definecolor{cvprblue}{rgb}{0.21,0.49,0.74}
\usepackage[pagebackref,breaklinks,colorlinks,allcolors=cvprblue]{hyperref}

\def\paperID{40} 
\def\confName{EarthVision}
\def\confYear{2026}

\title{Location Is All You Need: Continuous Spatiotemporal Neural Representations of Earth Observation Data}

\author{
Mojgan Madadikhaljan$^{1}$ \thanks{Corresponding author: mojgan.madadikhaljan@unibw.de}\quad
Jonathan Prexl$^{1}$ \quad
Isabelle Wittmann$^{2}$ \\
Conrad M. Albrecht$^{3,4}$ \quad
Michael Schmitt$^{1}$\\
$^{1}$University of the Bundeswehr Munich\quad
$^{2}$IBM Research -- Europe \\
$^{3}$Columbia University\quad
$^{4}$German Aerospace Center (DLR)\\[0.5em]
}
\begin{document}
\maketitle
\begin{abstract}
In this work, we present \textit{LIANet} (\textbf{L}ocation \textbf{I}s \textbf{A}ll You \textbf{N}eed N\textbf{et}work), a coordinate-based neural representation that models multi-temporal spaceborne \ac{EO} data for a given region of interest as a continuous spatiotemporal neural field. Given only spatial and temporal coordinates, \textit{LIANet} reconstructs the corresponding satellite imagery. Once pretrained, this neural representation can be adapted to various EO downstream tasks, such as semantic segmentation or pixel-wise regression, importantly, without requiring access to the original satellite data. 
\textit{LIANet} intends to serve as a user-friendly alternative to \acp{GFM} by eliminating the overhead of data access and preprocessing for end-users and enabling fine-tuning solely based on labels. We demonstrate the pretraining of \textit{LIANet} across target areas of varying sizes and show that fine-tuning it for downstream tasks achieves competitive performance compared to training from scratch or using established \acp{GFM}.
The source code and datasets are publicly available at \url{https://github.com/mojganmadadi/LIANet/tree/v1.0.1}.
\end{abstract}
    
\section{Introduction}
\label{sec:intro}

\ac{EO} data provide a dynamic and global record of our planet, forming the basis for many relevant applications such as climate monitoring, ecosystem management, agriculture, and disaster response \cite{kavvada2022earth,lang2023high,turkoglu2021crop}. The continuous growth of satellite constellations, sensor types, and the increase in revisit frequencies have led to a growth in both the volume and diversity of \ac{EO} data use cases \cite{cantrell2024earth, wilkinson2024environmental}. However, this increase in data volume and complexity makes the utilization of such data increasingly difficult for end-users, especially those from non-\ac{EO} related disciplines.
To tackle this challenge, the \ac{EO} community has started following the trend seen in language, vision, and multimodal deep learning research by transitioning toward the use of \acp{FM}. Here, self-supervised pretraining on large-scale datasets enables the models to learn general-purpose representations that can be efficiently adapted to a wide range of downstream tasks with minimal supervision \cite{uelwer2025survey, wang2022self}. This paradigm further empowers end-users to operate directly at the embedding level by utilizing pretrained encoder networks on mono-temporal, multi-temporal, or even multimodal \ac{EO} data \cite{jakubik2025terramind, brown2025alphaearth, xiong2024DOFA, jakubik2023prithvi,cong2022satmae,klemmer2025satclip,vivanco2023geoclip}.

\begin{figure}
    \centering
    \includegraphics[width=0.8\linewidth]{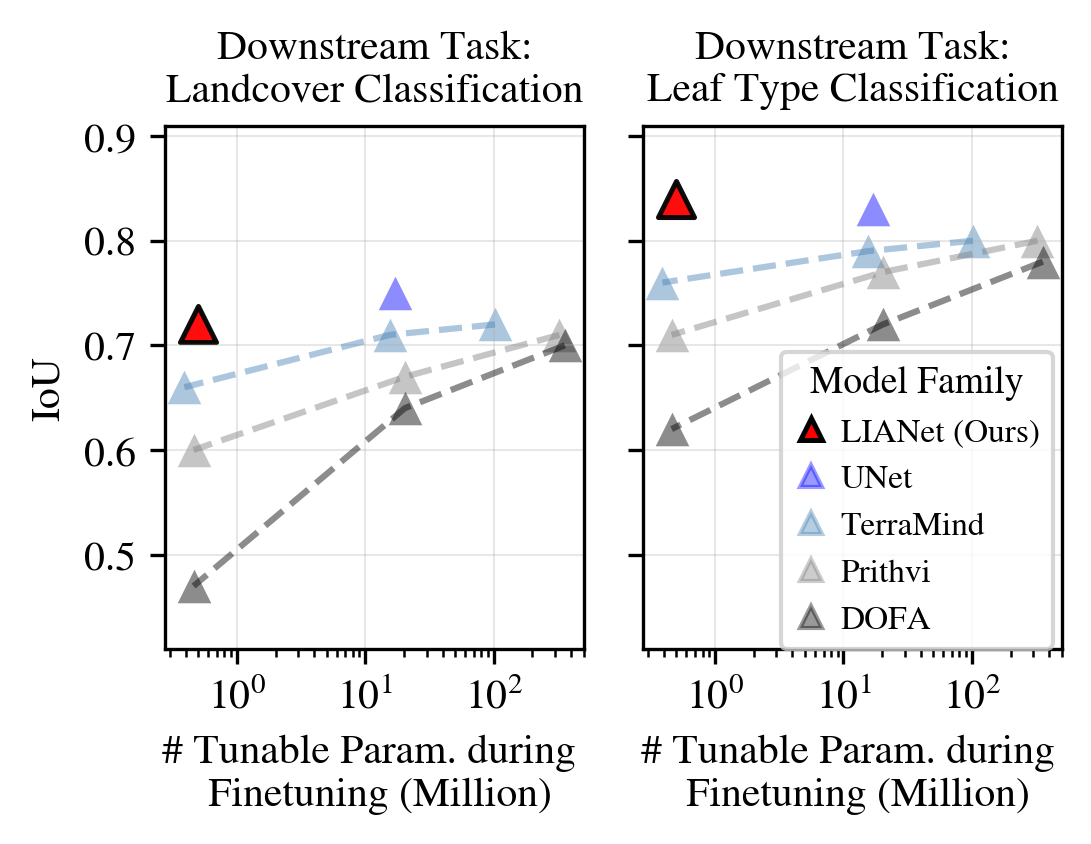}
    \caption{Comparison of model performance as a function of the number of tunable parameters for two pixel-wise classification tasks. A classical UNet architecture (trained from scratch), as well as three commonly applied \acp{FM} (with varying parameter count, depending on fine-tuning scheme), serve as a baseline. Our approach, \textit{LIANet}, achieves a high performance with minimal tunable parameters.}
    \label{fig:introplot}
\end{figure}

In this work, we propose a new paradigm for EO representation learning: a coordinate-based neural representation that directly models a specific geographic area (in this study, up to $\approx$ \SI{12000}{\square\kilo\metre}) as a latent grid inspired by \acp{INR} research \cite{muller2022instant,essakine2024we}. 
In our setup, the model input solely consists of spatio-temporal coordinates $(x,y,t)$, which are mapped to a representation $\hat{\textbf{g}}_{x,y,t}$ at the corresponding grid position. A decoder network $D$ is trained such that it generates the corresponding satellite image patch $\mathbf{I}$ at the time of interest. We refer to this generative training procedure $\mathbf{I}=D(\hat{\textbf{g}}_{x,y,t})$ as \textit{pretraining} in the remainder of the manuscript, where details can be found in \cref{sec:method}.
Once pretrained, this neural representation of the Earth’s surface parametrized by $(x,y,t)$ can then be used for data inspection and fine-tuned (by adapting $D$) for arbitrary downstream tasks, as will be shown in \cref{sec:experimentalsetup}.

Our approach offers three main advantages:
Firstly, during fine-tuning, no additional satellite data is required, since the relevant spatiotemporal information has been encoded during the generative pretraining. From an end-user perspective, this enables a lightweight adaptation workflow for interacting with \ac{EO} data: given our pretrained model for a region of interest (\eg, a country), the user can adapt it to any target task using only a small set of labels, without the need to access the raw data or preprocessing. This paradigm is similar to embedding-based workflows, where representations are precomputed and reused across tasks. However, unlike embedding approaches, our approach preserves direct access to the underlying image 
data, allowing users to view the area at any point in time.
Secondly, the continuous-space/time generative objective provides an intrinsic, label-agnostic mechanism to assess the quality of the learned representations as well as temporal change monitoring within the model itself. As with other self-supervised learning approaches, reconstruction performance in \textit{LIANet} provides a direct, quantitative measure independent of downstream tasks. This complements standard evaluation protocols, which rely on task-specific benchmarks and are often influenced by dataset characteristics such as label noise, class imbalance, or benchmark biases.
Lastly, the design of the multi-resolution latent grid inherently encourages the model to learn spatially continuous representations, reflecting the natural continuity of the Earth’s surface (see \cref{sec:method}), in contrast to more common strategies that generate patch-wise embeddings (see \cref{sec:sota}).

It is important to emphasize that one key difference from classical representation learning lies in the \textit{hyper-local} nature of our approach. While traditional pretrained networks are designed for broad global applicability, our model is intentionally designed to produce meaningful embeddings within the region used during pretraining. The \textit{hyper-local} design is particularly suitable for scenarios in which an end user, for example, a governmental agency or regional authority, focuses exclusively on environmental monitoring or sustainable development within a specific area of interest, such as a municipality or a single country. In such cases, global generalization is often secondary to obtaining a highly performant, region-optimized model that can be efficiently adapted to multiple downstream tasks. Therefore, this workflow is most practical when embeddings are generated centrally, for instance by space agencies or other large-scale data providers, and subsequently distributed to regional stakeholders. As demonstrated throughout this manuscript, the advantages of our approach, including interpretability, strong downstream performance, and parameter efficiency, yield substantial benefits.

Overall, our workflow and corresponding contributions can be summarized as follows:
\begin{enumerate}
    \item We pretrain an INR-based network mapping spatio-temporal coordinates to neural representations. Our approach builds on a discrete grid of embedding vectors to model a continuous field of representations that we decode into corresponding \ac{EO} data.
    We will demonstrate the general capabilities by utilizing multi-temporal observations of the multi-spectral satellite \textit{Sentinel-2}. Nevertheless, more general approaches using multi-satellite data extensions can be implemented similarly. 
    \item We evaluate the reconstruction quality on multi-temporal observations and demonstrate how the learned representation can be applied for visual inspection.
    \item We utilize the pretrained models to perform dense few-shot fine-tuning on five common downstream tasks (two pixel-wise regression and three semantic segmentation tasks) in the main paper, and additionally evaluate on two adapted datasets from the PANGAEA benchmark \cite{marsocci2024pangaea} in the supplementary material. We compare the results against differently sized UNets trained from scratch and three widely used \acp{GFM} as baselines. An overview of the results, contextualized by the number of tunable parameters during fine-tuning, is provided in \cref{fig:introplot} for two representative downstream applications.
\end{enumerate}

\section{Related Works}
\label{sec:sota}

We place LIANet in the context of prior work on \acp{GFM}, precomputed embeddings, and coordinate-based models, emphasizing the key differences outlined in \cref{tab:comparison}.

\paragraph{\Acp{GFM}.} Recent work has introduced large-scale \acp{GFM} trained on single-modal \cite{dumeur2024ubran,yu2024metaearth,klemmer2025satclip,braham2025spectralearth,prexl2024senpamaesensorparameteraware,prexl2025sar} and multi-modal \ac{EO} imagery \cite{szwarcman2024prithvi,jakubik2023prithvi,cong2022satmae,xiong2024DOFA,fuller2023croma,bi2025ringmoe,velazquez2025earthview,guo2024skysense,hong2023spectralgpt,prexl2023multi}, as well as on vision–language datasets \cite{jakubik2025terramind,khanna2023diffusionsat,liu2024remoteclip,li2023rs,brown2025alphaearth}. Most adopt masked autoencoding (MAE) \cite{jakubik2025terramind,jakubik2023prithvi,schmude2024prithviwxc,cong2022satmae,xiong2024DOFA,velazquez2025earthview,hong2023spectralgpt,prexl2024senpamaesensorparameteraware,prexl2025sar} or contrastive learning \cite{klemmer2025satclip,liu2024remoteclip,li2023rs,guo2024skysense,brown2025alphaearth,prexl2023multi} as primary pretraining objectives. Others combine both paradigms \cite{fuller2023croma}, or utilize a more complex framework, such as mixture-of-experts \cite{bi2025ringmoe}. While transformer-based architectures dominate among current \acp{GFM}, some integrate convolutional or hybrid CNN–ViT designs to improve computational efficiency~\cite{klemmer2025satclip,liu2024remoteclip,li2023rs}. These models achieve strong performance across diverse downstream tasks, including spatially dense prediction. However, they require raw imagery and substantial compute at inference. We benchmark against three representative \acp{GFM}: TerraMind \cite{jakubik2025terramind}, an any-to-any generative multimodal model, DOFA \cite{xiong2024DOFA}, which adapts a single transformer across sensor types, and Prithvi-EO-2.0 \cite{szwarcman2024prithvi}, providing large parameter count and spatiotemporal embeddings.

\paragraph{Precomputed Embeddings and Location Encoders.}
Precomputed embedding products provide aggregated spatiotemporal representations derived from foundation models, enabling querying at fixed spatial and temporal resolutions without requiring raw imagery~\cite{brown2025alphaearth, feng2025tessera}. However, they trade flexibility for efficiency: representations are temporally aggregated, do not support continuous interpolation, and do not allow for image reconstruction. Coordinate-based encoders map geographic locations directly to semantic features, further reducing data requirements, but remain primarily discriminative and coarse in spatial detail~\cite{klemmer2025satclip, vivanco2023geoclip}.

\paragraph{Implicit Neural Representations and Generative Models.} A growing line of research focuses on generative models for \ac{EO} image synthesis based on text or paired image inputs \cite{liu2024diffusionsurvey}. These include diffusion-based models designed for synthetic image generation, cross-modal translation, and scene rendering, such as \cite{yu2024metaearth,liu2025text2earth,sastry2024geosynth,sebaq2024rsdiff,tang2024crs,khanna2023diffusionsat}. Despite producing visually realistic and compelling outputs, their main goal is not to reflect the exact real-world content of a given location but rather the generation of synthetic data. \Acp{INR} model signals as continuous functions and have recently been explored as an effective approach for image compression \cite{gomes2025lossy,sitzmann2020implicit,dupont2021coin,strumpler2022implicit}, by learning compact implicit representations of visual scenes. Building on this foundation, several studies have extended INR-based compression to \ac{EO} imagery \cite{li2023remote_,rezasoltani2024hyperspectral_,zhang2024compressing_,cho2024neural_}. Building on this line of research, we extend INR-based modeling to multi-temporal reconstruction, patch-level decoding, and efficient representation of larger spatial areas.

\begin{table}[t!]
\centering
\footnotesize
\setlength{\tabcolsep}{4pt}
\renewcommand{\arraystretch}{1.05}
\caption{Conceptual comparison of \acp{GFM}, precomputed embeddings, location encoders, and LIANet on key modeling axes.}
\label{tab:comparison}
\begin{tabular}{p{2.35cm}cccc}
\toprule
&
\thead{\acp{GFM}} &
\thead{Precomputed\\Embeddings} &
\thead{Location\\Encoders} &
\textbf{\thead{LIANet\\(ours)}} \\
\midrule

No raw imagery\\
at inference
& \xmark & \cmark & \cmark & \cmark \\

Continuous spatio-\\
temporal field
& \xmark & \xmark & Spatial only & \cmark \\

Image\\
reconstruction
& \cmark & \xmark & \xmark & \cmark \\

Dense, high-res\\
downstream tasks
& \cmark & Limited & \xmark & \cmark \\

Hyper-local\\
spatial detail
& \xmark & \xmark & \xmark & \cmark \\

\bottomrule
\end{tabular}
\end{table}
\section{Methodology}
\label{sec:method}

\begin{figure*}[ht!]
   \centering
   \includegraphics[width=0.86\textwidth]{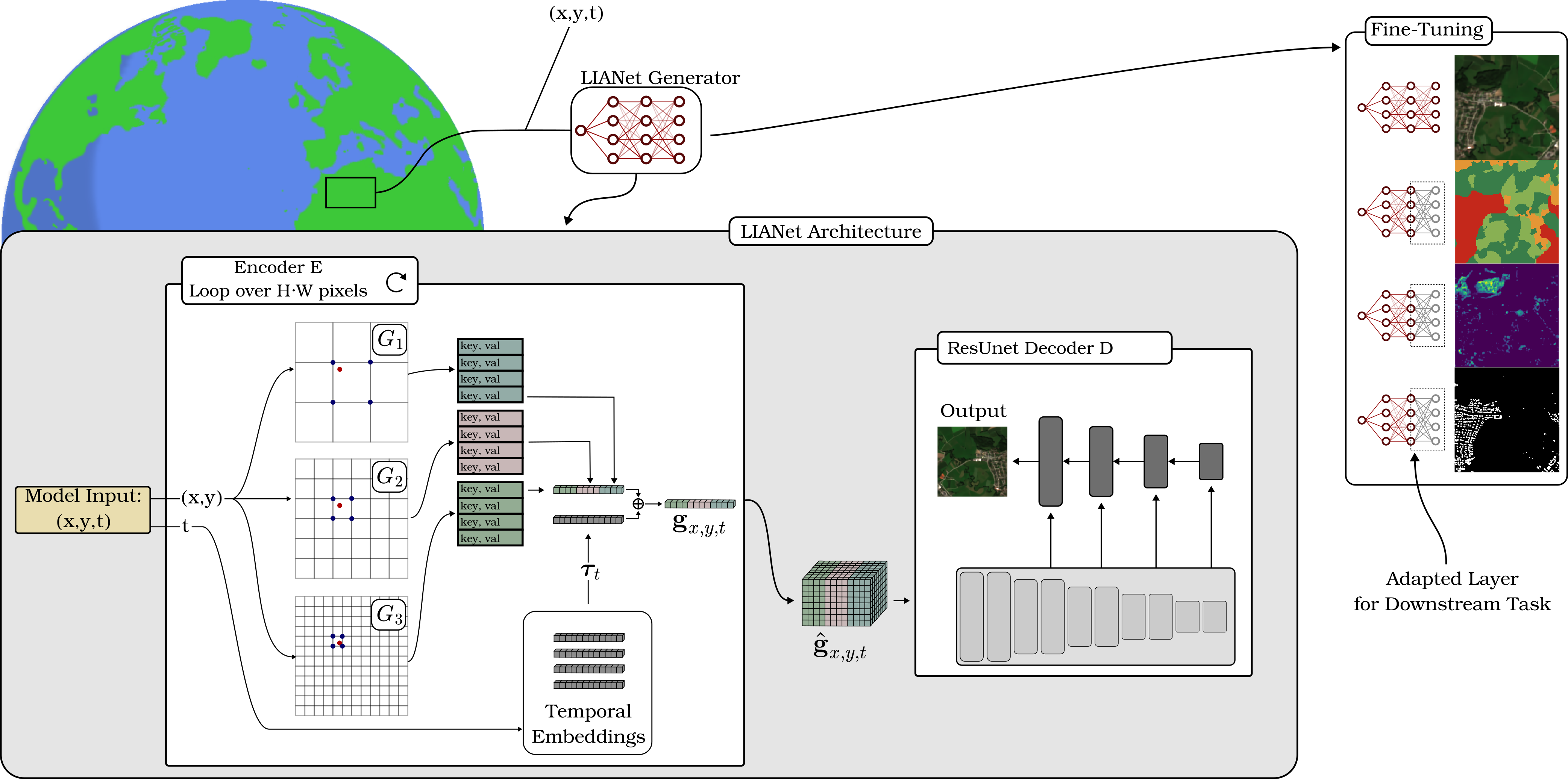}
    \caption{Visualization of the \textit{LIANet} workflow. \textbf{Pretraining:} The specified $(x, y)$ coordinates (model input) are used to query learned representations from a hash table by interpolating table entries for the neighboring pixels (blue dots) for the given input position (red dots) for all different resulution grids $G_i$ (in this illustration $N_\text{grids}=3$) and adding them to the corresponding temporal encoding vector. This process is applied iteratively to all pixels of the target output image and is implemented in a computationally efficient manner \cite{muller2022instant}, leading to a scene representation $\hat{\textbf{g}}_{x,y,t}$. A decoder network $D$ reconstructs the underlying satellite image $\textbf{I}=D(\hat{\textbf{g}}_{x,y,t})$. \textbf{Fine tuning:} In this stage, the final layers of $D$ are adapted to generate labels for an arbitrary downstream task. During fine-tuning, the learned table values and the first few decoder layers are frozen, and no input images are required during this stage.}
   \label{fig:concept}
\end{figure*}

This section outlines the design of the network architecture employed to encode spaceborne \ac{EO} data for a given region, along with the procedure for fine-tuning the pretrained model to an arbitrary downstream task. Note that the terms \textit{encoder} and \textit{decoder} are used for clarity and do not refer to a classical bottlenecked architecture. Instead, the representation is distributed across the grid parameters and the CNN, with the full model jointly encoding the region.

\paragraph{Encoder Stage.}
In order to encode a target area $\mathcal{A}$, we first define multiple grids $G_i$ with $i \in \{1,\dots, N_{\text{grids}}\}$ at multiple resolution levels $L_{i}$.
Inspired by recent advances in Neural Radiance Fields~(NeRFs)~\cite{muller2022instant}, which offer high computational efficiency, each of the grids has a corresponding table that stores learnable embeddings with fixed length $T$ (number of table entries) and embedding dimension of $F$. Every node of the grid is assigned to an index in the table via a hashing operation (for a detailed description see~\cite{muller2022instant}). The design choices regarding the number of grids, as well as their respective grid spacings, are adapted to the specific characteristics of the EO data used in this work and will be described in detail in the following section.
For reference, \cref{fig:concept} provides a graphical overview of the workflow.
For a given point $(x,y)$ in the target area $\mathcal{A}$, we query the indices of the four surrounding nodes from their corresponding table at all resolution levels. This results in four feature vectors per grid $G_i$. The feature vector at query point $(x,y)$ is obtained from feature-wise bilinear interpolation given the four grid feature vectors.
This interpolation step forces the representations to be continuous in space, where spatially close points (depending on the specific level $L_{i}$) are assigned to similar representations, reflecting the continuous nature of EO data. 
The results for each grid $G_i$ get concatenated to an embedding for the position of the overall dimension $\mathbb{R}^{N_{\text{grids}} \cdot F}$ and summed up with a learnable temporal encoding vectors $\boldsymbol{\tau}_t$ for each acquisition time step $t$ in the dataset, where each $\boldsymbol{\tau}_t$ is represented as vectors in $\boldsymbol{\tau}_t \in \mathbb{R}^{{N_{grids}} \cdot F}$.
We refer to the above steps as the encoder $E$ part of the model, 
\[
E:\mathbb{R}^3 \to \mathbb{R}^{{N_\text{grids}} \cdot F} \text{ ,}
\]
for which the output for one specific position $(x,y,t)$ will be denoted as $\textbf{g}_{x,y,t} \in \mathbb{R}^{{N_\text{grids}} \cdot F}$.
Since the objective during pretraining is to reconstruct the corresponding satellite image $\textbf{I} \in \mathbb{R}^{C \times W \times H}$ ($C$: channels, $W$: width, and $H$: height), the encoder $E$ is queried for all pixels in the corresponding image patch
$(x+\delta_x,y+\delta_y)$ with 
$\delta_x \in \{0,\dots, W\}$
and
$\delta_y \in \{0,\dots, H\}$\footnote{The steps $(\delta_x, \delta_y)$ are directly determined by the ground sampling distance of the satellite imagery, described in \cref{sec:experimentalsetup}.}
iteratively, resulting in multiple $\textbf{g}_{x,y,t}$ (one per pixel in the output image) that we reassemble into the image format and denote as $\hat{\textbf{g}}_{x,y,t} \in \mathbb{R}^{{N_\text{grids}} \cdot F \times W \times H}$, which encodes the full area of the corresponding satellite image and is passed to a decoder network, described in the following section. 

\paragraph{Decoder Stage.}
The resulting latent representation $\hat{\textbf{g}}_{x,y,t}$ is passed to a CNN-based decoder $D$,
\[D:\mathbb{R}^{{N_\text{grids}} \cdot F \times W \times H} \to \mathbb{R}^{C\times W \times H}\text{ ,}
\]
which follows a \textit{ResNet–UNet hybrid} design~\cite{qubvel_segmentation_models_pytorch}. 
The decoder outputs the reconstructed satellite image $\textbf{I} \in \mathbb{R}^{C\times H \times W}$ at position $(x, y)$ and time $t$ and thus can be pretrained in a self-supervised fashion by reconstructing the corresponding image.
It is important to point out that, unlike typical INR designs that render by repeatedly querying a fully-connected MLP, our method generates $\mathbf{I}$ with a single forward pass of $D$. This directly exposes the final layers of $D$ as a modifiable module that can be easily adapted to downstream tasks (\eg, semantic segmentation or pixel-wise regression).
The overall process is summarized in \cref{alg:LIANetInference}.

\paragraph{Pretraining and Fine-tuning.}
During pretraining, random spatial points within the area of interest $\mathcal{A}$ are sampled in each epoch to ensure spatial continuity in the learned latent representations. 
Given the model input $(x, y, t)$, the decoder output is trained to match the corresponding satellite image using an $L_1$ loss function.

For adapting the pretrained model to an arbitrary downstream task, we freeze the encoder $E$ containing the corresponding learnable table entries as well as the initial layers of the decoder $D$ and modify the final layers of the decoder $D$ to match the output format of the specific downstream application. 
In this fine-tuning stage, no satellite imagery is required, and only $(x,y,t)$ are the inputs to the model. The fine-tuning can be performed in a few-shot manner using only a small number of labeled samples covering a limited subset of the area $\mathcal{A}$, as will be demonstrated in \cref{sec:results}.
For the remainder of the manuscript, we will refer to the proposed model as \textbf{L}ocation \textbf{I}s \textbf{A}ll you \textbf{N}eed N\textbf{et}work (\textbf{LIANet}).

\begin{algorithm}[H]
\caption{\textit{LIANet} Forward Pass}
\label{alg:LIANetInference}
\footnotesize
\begin{algorithmic}[1]

\Require Encoder $E$, Decoder $D$, $(x,y)$, $(\delta_x,\delta_y)$, $t$

\vspace{4pt} 

\State Initialize embedding $\hat{\mathbf{g}}_{x,y,t} \in \mathbb{R}^{N_\text{grids}\cdot F \times W \times H}$
\For{$i = 0$ to $W-1$}
  \For{$j = 0$ to $H-1$}
    \State $\hat{\mathbf{g}}_{x,y,t}[:,i,j] \gets E(x+i\delta_x,\ y+j\delta_y,\ t)$
  \EndFor
\EndFor

\vspace{2pt}

\State $\mathbf{I} \gets D(\hat{\mathbf{g}}_{x,y,t})$
\Comment{decoder produces final image}
\end{algorithmic}
\end{algorithm}
\section{Experimental Setup}
\label{sec:experimentalsetup}

In this section, we present the details of the \pretraining and \finetuning stages. In addition, we describe the implementation details of the corresponding reference models used later in \cref{sec:results}.

\subsection{Pretraining and Model Settings}

Within this study, we consider three area $\mathcal{A}$ sizes, denoted as $\mathcal{A}_{0}$, $\mathcal{A}_{+}$ and $\mathcal{A}_{++}$ for which multispectral satellite data of the \textit{Sentinel-2} mission~\cite{drusch2012sentinel} data is acquired at four time points at different seasons from Munich, Germany. The areas are overlapping in the sense that larger areas are an extension of the $\mathcal{A}_{0}$, \ie $\mathcal{A}_0\subset\mathcal{A}_+\subset\mathcal{A}_{++}$, where $\mathcal{A}_0$ covers $\SI{2500}km^2$, $\mathcal{A}_+$ covers $\SI{5000}km^2$ and $\mathcal{A}_{++}$ covers $\SI{12000}km^2$. 

Generative \pretraining is then conducted by collecting about a million random samples $\textbf{I} \in \mathbb{R}^{C \times W \times H}$ per epoch. We grow the number of pretraining epochs linearly with the area of the encoded target area, and two different sizes of the decoder network $D$ are tested and referred to as $D_{base}$ and $D_{large}$ (or \textit{LIANet-Base} and \textit{LIANet-Large} if referred to the complete model setup) with $75\,$M and $133\,$M trainable parameters, respectively. The specific number of epochs, as well as all other training settings, can be found in the \textbf{supplementary materials}. 

\paragraph{Grid Size Settings.} All \textit{Sentinel-2} channels are resampled to a uniform Ground Sampling Distance (GSD) of 10 meters. We fix the patch dimensions to $H=W=128$ for all experiments. Accordingly, the grid dimensions $G_i$ are defined as follows:
The coarsest grid ($G_1$), with a node spacing of $6{,}800$ meters, contains multiple $128\times128$ image patches. We then define ten additional grids ($N_{\text{grids}}=11$ in total), each with the node spacing reduced by a factor of two relative to the previous level. The finest grid ($G_{11}$) thus achieves a node distance below 10 meters, enabling the encoding of sub-pixel level information. The nested grids are depicted in \cref{fig:concept}.

The corresponding hash tables that store the learnable latent representations have a fixed size of $T = 2^{17}$ entries with an embedding dimension of $F = 4$. Hence, querying concatenated embeddings from all $11$~grid resolutions leads to an overall representation $\hat{\textbf{g}}_{x,y,t}$ of the shape~$\mathbb{R}^{44 \times 128 \times 128}$.
This serves as input to CNN-based decoder $D$ that generates image patch $\textbf{I}$, where $\textbf{I}=D(\hat{\textbf{g}}_{x,y,t}) \in \mathbb{R}^{12\times 128 \times 128}$ with $C=12$ for \textit{Sentinel-2}.

\begin{figure*}[t!]
    \centering
    \includegraphics[width=0.86\textwidth]{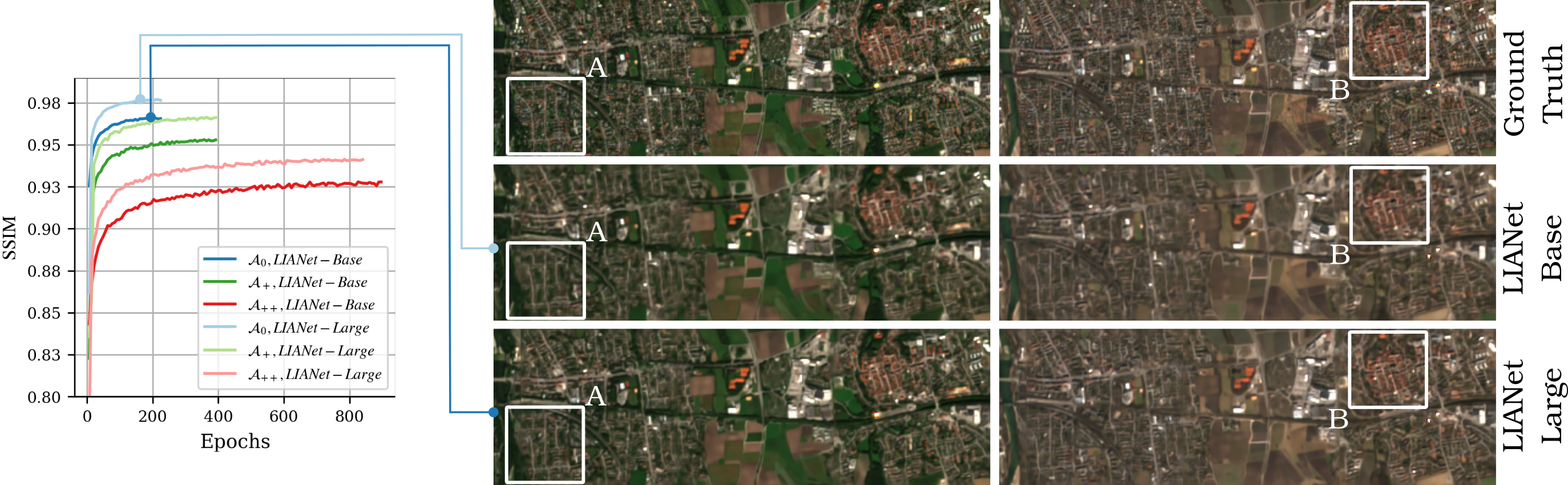}
    \caption{Left plot shows the reconstruction performance of \textit{LIANet-Base} and \textit{LIANet-Large} on $\mathcal{A}_{0}$, $\mathcal{A}_{+}$, and $\mathcal{A}_{++}$ measured by Structural Similarity (SSIM) of the reconstructed images as a function of pretraining step. The images on the right show a sample ground truth image (first row), reconstruction images of \textit{LIANet-Base} (second row) and \textit{LIANet-Large} (third row) from two different seasons. 
    The enhanced reconstruction capability of \textit{LIANet-Large} compared to \textit{LIANet-Base} becomes evident in regions with high-frequency features, such as the highlighted areas $A$ and $B$.
    Note that the images produced on the right are the result of four adjacent and non-overlapping $128 \times 128$ patches stitched together horizontally, where, due to effective continuous reconstruction, no patching effect appears, enabling larger scene reconstructions without any undesired artifacts.}
    \label{fig:reconstruction_visual}
\end{figure*}

\subsection{Fine-tuning Settings}

To evaluate the generalization capability of our approach, we fine-tune the pretrained models on five diverse downstream applications. It is important to note that, for a proper evaluation, corresponding labels must be available within the encoded area $\mathcal{A}$. Many existing \ac{EO} benchmarks are not directly compatible with our framework, as they (i) lack georeferencing, (ii) provide a very small number of samples within a Sentinel-2 tile, (iii) rely on input modalities differing from multispectral Sentinel-2 imagery, or (iv) are not fully open-access for large-scale generative pretraining. We therefore construct our custom datasets to ensure geographically contiguous coverage consistent with our coordinate-based design. Beyond this contiguous setup, we additionally evaluate \textit{LIANet} on adapted versions of two standardized PANGAEA benchmark datasets, namely PASTIS~\cite{pastis_garnot2021panoptic} and HLS Burn Scars~\cite{HLS_Foundation_2023}. Due to space constraints, detailed protocols and results are reported in \cref{sec:pastis_burnscars} of \textbf{supplementary material}.
The custom dataset is generated using common EO applications \cite{MBF,brown2022dynamic,clms_dlt_2018_present,meta_wri_chm_2024}. 

To create the dataset, we pair the \textit{Sentinel-2} data with labels for land-cover classification \cite{brown2022dynamic} (six different land cover classes in our target area provided for every season), canopy height regression \cite{meta_wri_chm_2024}, dominant leaf type classification \cite{clms_dlt_2018_present} (two leaf type classes plus background) as well as data for building footprints \cite{MBF} which will be used in a regression type setup (by predicting pixel-wise percentage of building coverage as suggested in \cite{fibaek2024phileo}) as well as a binary segmentation task. It should be noted that for the building footprint semantic segmentation task, we employ further upsampling layers to predict the building mask on a $2.5\,$m scale (tighter than the native resolution of the sensor) as done in \cite{prexl2023potential,prexl2024comparison}. Since the baselines described in the next paragraph have no native capabilities to predict beyond the native sensor resolution, we will use a limited set of baselines for this task only. It is to be mentioned that only land cover classes \cite{brown2022dynamic} are provided for each timestep individually, whereas the rest are available as a single label for all temporal spans.

A limited portion of $\mathcal{A}_{0}$ (\SI{500}{\square\kilo\metre}) is used for fine-tuning, corresponding to roughly $20\%$, $10\%$, and $4\%$ of $\mathcal{A}_{0}$, $\mathcal{A}_{+}$, and $\mathcal{A}_{++}$, respectively, whereas the remaining area is used for validation. We will additionally provide further ablation experiments regarding the amount (area) of labeled data and the corresponding model performance. 

In order to adapt the model output to the corresponding downstream task, we drop the last convolutional layer of the decoding network $D$ and introduce six trainable convolutional layers (the rest of $D$ as well as all trainable parameters in $E$ are frozen), leading to a parameter-efficient fine-tuning setting with only $0.5\,$M trainable parameters.

The corresponding loss functions, learning rates, and number of epochs can be found in \cref{sec:Suppmat_finetining_details} of  \textbf{supplementary materials}. 
The used \textit{Sentinel-2} images as well as all corresponding labels will be available together with the full code for model pretraining and fine-tuning within the project repository. 

\subsection{Baselines}
We compare the performance of our models to two different-sized task-specific UNet baselines (trained from scratch) and three widely used \acp{GFM}, namely: \textit{TerraMind-Base}, \textit{Prithvi v2-300}, and \textit{DOFA-Large}. Each \ac{FM} is adapted to the downstream task by coupling its corresponding encoder with a task-specific decoder~\cite{gomes2025terratorch}. We evaluate three different fine-tuning settings:
\begin{itemize}
    \item \textit{Full fine-tuning:} both encoder and decoder parameters are updated.  
    \item \textit{Frozen-backbone fine-tuning:} the encoder remains frozen and only the decoder is trained.  
    \item \textit{Embedding setup:} only the final-layer embeddings are extracted from the backbone, and a lightweight fully-convolutional (FCN) decoder is trained on top.
\end{itemize}
Both the \textit{full} and \textit{frozen} setups employ a standard UNet decoder that operates on four intermediate and final feature representations, while the embedding setup uses only the last-layer embedding and a compact FCN decoder. Consequently, the embedding configuration has the fewest trainable parameters (below 0.5~M) and represents a practical embedding workflow, where final-layer embeddings can be extracted once and shared as data representations.

All \acp{FM} operate on multispectral optical inputs. Specifically, \textit{TerraMind-Base} and \textit{DOFA-Large (ViT-Patch16-224)} use all available \textit{Sentinel-2} bands, while \textit{Prithvi v2-300} uses the subset of \textit{Sentinel-2} bands, matching the HLS \cite{szwarcman2024prithvi} (original pretraining data distribution) bands, as input. The corresponding loss functions, learning rates, and the number of epochs for fine-tuning all corresponding benchmark \acp{FM} can be found in the \cref{sec:Suppmat_finetining_details} of \textbf{supplementary materials}. 
\section{Results}
\label{sec:results}

This section details our empirical findings. We begin by evaluating reconstruction quality and report the performance of adapted pretrainings to several downstream tasks. Finally, we present the benchmarks and experiments used to investigate the label efficiency of our approach.

\paragraph{Pretraining.}
\Cref{fig:reconstruction_visual} provides a visual overview of the model’s generative capabilities when tasked with reconstructing the underlying data at a specific geographic location $(x,y,t)$ for a range of acquisition times $t$ (different seasons). The figure compares results obtained using two decoder configurations of different capacity, namely $D_{base}$ and $D_{large}$, where the latter results in a sharper appearing reconstruction for high-frequency features (compare marked areas A and B in \cref{fig:reconstruction_visual}). The generated samples also show that the temporal embeddings are effective in capturing the changes across different seasons. In addition to the qualitative reconstructions, \cref{fig:reconstruction_visual} also reports the corresponding structural similarity index (SSIM) as a function of the pretraining epochs.
Overall, the results highlight two key factors that influence reconstruction quality during pretraining: the model size and the spatial extent of the encoded area $\mathcal{A}$. Increasing the decoder size consistently improves reconstruction quality, while conversely, enlarging the encoded spatial area leads to a degradation in reconstruction quality. An extended analysis with respect to reconstruction quality investigated with tools from the image compression literature can be found in the \cref{sec:neural_compression} of \textbf{supplementary materials}.

\paragraph{Downstream Task Performance.}
Before discussing the results, it is important to clarify the scope of the proposed method. \textit{LIANet} adopts a hyper-local design, focusing solely on a target area rather than generalizing to unseen regions, which is the main objective of \acp{GFM}. Accordingly, the benchmarking results should be interpreted in this context: \textit{LIANet} is region-specific and provides an alternative \ac{EO} representation learning approach, whereas \acp{FM} target global applicability. These experiments aim to show that, within its intended setting, LIANet achieves strong performance compared to state-of-the-art methods.

The results for adapting the pretrained \textit{LIANet} models to the five downstream tasks can be found in \cref{tab:results_5k} (for the three pixel-wise classification tasks and the two regression tasks), together with the three FM baselines as well as from-scratch-training of the \textit{UNet} \cite{jiangtao2025comprehensive} and \textit{Micro UNet} with reduced tunable parameters.
\cref{tab:results_5k} contains the results of the experiments on $\mathcal{A}_{0}$. For results on $\mathcal{A}_{+}$ and $\mathcal{A}_{++}$, as well as visual examples, see \cref{sec:Suppmat_finetining_details} in \textbf{supplementary materials}. 

Comparing the results of \cref{tab:results_5k}, \textit{LIANet} consistently performs within the top three approaches across all metrics (for both $D_{base}$ and $D_{large}$), even with the relatively small tunable parameter count of $0.5\,$M in the fine-tuning. Comparing this to the \acp{FM}, one can observe a significant performance gain of the \textit{hyper-local} approach of \textit{LIANet} when choosing a comparable parameter count (\textit{Embedding} configuration) and even a performance gain when comparing against the fully fine-tuned ($\approx 100\,$M parameters) setup.

\begin{table*}[t!]
    \centering
    \footnotesize
    \caption{(Top) Pixel-wise classification task performance measured with Intersection-over-Union (IoU), Accuracy (Acc), and F1-score (all calculated with macro averaging) along with the tunable parameter counts for the area of interest $\mathcal{A}_0$.
    (Bottom) Regression task performance (Mean Absolute Error (MAE) and Mean Squared Error (MSE)) along with the tunable parameter counts for the area of interest $\mathcal{A}_0$. The from-scratch trainings and \acp{FM} are evaluated in different configurations with a varying number of tunable parameters.
    For all metrics reported, the corresponding \textbf{top three} performances are printed \textbf{bold}. Due to the limited number of comparisons, we are not marking any models for the task of building footprint segmentation.}
\resizebox{\textwidth}{!}{
    \begin{tabular}{clcccc}
    \toprule
    \textbf{Task} & \textbf{Model / Setting} & \textbf{\# Tunable Params (M)} & \textbf{IoU} & \textbf{ACC} & \textbf{F1} \\
    \midrule

    \multirow{7}{*}{\makecell{Dynamic \\ World}}
      & UNet\,/\,Micro UNet  & 17.3\,/\,0.49 & \textbf{0.75}\,/\,0.66 & \textbf{0.82}\,/\,0.73 & \textbf{0.84}\,/\,0.74 \\
      & TerraMind-base (Full\,/\,Frozen\,/\,Embedding) & 102\,/\,15.5\,/\,0.39 & \textbf{0.72}\,/\,\textbf{0.71}\,/\,0.66 & \textbf{0.78}\,/\,0.77\,/\,0.73 & \textbf{0.81}\,/\,\textbf{0.80}\,/\,0.73 \\
      & Prithvi v2-300 (Full\,/\,Frozen\,/\,Embedding) & 324\,/\,20.3\,/\,0.46 & \textbf{0.71}\,/\,0.67\,/\,0.60 & 0.77\,/\,0.74\,/\,0.68 & \textbf{0.80}\,/\,0.76\,/\,0.69 \\
      & DOFA-Large (Full\,/\,Frozen\,/\,Embedding)     & 357\,/\,20.3\,/\,0.46 & 0.70\,/\,0.64\,/\,0.47 & 0.76\,/\,0.71\,/\,0.54 & 0.79\,/\,0.74\,/\,0.57 \\
      & LIANet-Base & 0.5 & \textbf{0.72} & \textbf{0.82} & \textbf{0.81} \\
      & LIANet-Large & 0.5 & \textbf{0.72} & \textbf{0.81} & \textbf{0.81} \\
    \arrayrulecolor{black!30}\cmidrule(lr){1-6}

    \multirow{7}{*}{\makecell{Dominant \\ Leaf Type}}
      & UNet\,/\,Micro UNet & 17.3\,/\,0.49 & \textbf{0.83}\,/\,0.79 & \textbf{0.90}\,/\,0.87 & \textbf{0.90}\,/\,0.88 \\
      & TerraMind-base (Full\,/\,Frozen\,/\,Embedding) & 102\,/\,15.5\,/\,0.39 & \textbf{0.80}\,/\,0.79\,/\,0.76 & \textbf{0.88}\,/\,0.86\,/\,0.85 & \textbf{0.89}\,/\,0.87\,/\,0.86 \\
      & Prithvi v2-300 (Full\,/\,Frozen\,/\,Embedding) & 324\,/\,20.3\,/\,0.46 & \textbf{0.80}\,/\,0.77\,/\,0.71 & \textbf{0.88}\,/\,0.85\,/\,0.81 & \textbf{0.89}\,/\,0.86\,/\,0.82 \\
      & DOFA-Large (Full\,/\,Frozen\,/\,Embedding)     & 357\,/\,20.3\,/\,0.46 & 0.78\,/\,0.72\,/\,0.62 & 0.86\,/\,0.81\,/\,0.73 & 0.87\,/\,0.83\,/\,0.74 \\
      & LIANet-Base & 0.5 & \textbf{0.84} & \textbf{0.91} & \textbf{0.91} \\
      & LIANet-Large & 0.5 & \textbf{0.84} & \textbf{0.91} & \textbf{0.91} \\
    \arrayrulecolor{black!30}\cmidrule(lr){1-6}

    \multirow{3}{*}{\makecell{Building Footprint \\ Segmentation}}
      & UNet\,/\,Micro UNet & 17.3\,/\,0.49 & 0.76\,/\,0.69 & 0.87\,/\,0.76 & 0.85\,/\,0.78 \\
      & LIANet-Base & 0.5 & 0.68 & 0.77 & 0.77 \\
      & LIANet-Large & 0.5 & 0.70 & 0.81 & 0.79 \\

    \arrayrulecolor{black!100}\toprule
    \textbf{Task} & \textbf{Model / Setting} & \textbf{\# Tunable Params (M)} & \textbf{MAE} & \textbf{MSE} &  \\
    \midrule

    \multirow{6}{*}{\makecell{Canopy \\ Height}}
      & UNet\,/\,Micro UNet & 17.3\,/\,0.49 & \textbf{0.053}\,/\,0.072 & \textbf{0.013}\,/\,0.023 \\
      & TerraMind-base (Full\,/\,Frozen\,/\,Embedding) & 102\,/\,15.5\,/\,0.39 & \textbf{0.051}\,/\,\textbf{0.053}\,/\,0.113 & \textbf{0.013}\,/\,\textbf{0.013}\,/\,0.056 & \\
      & Prithvi v2-300 (Full\,/\,Frozen\,/\,Embedding) & 324\,/\,20.3\,/\,0.46 & \textbf{0.051}\,/\,0.056\,/\,0.113 & \textbf{0.012}\,/\,0.015\,/\,0.056  & \\
      & DOFA-Large (Full\,/\,Frozen\,/\,Embedding)     & 357\,/\,20.3\,/\,0.46 & 0.055\,/\,0.061\,/\,0.113 & \textbf{0.014}\,/\,0.018\,/\,0.056  & \\
      & LIANet-Base & 0.5 & \textbf{0.052} & \textbf{0.013}  & \\
      & LIANet-Large & 0.5 & 0.055 & \textbf{0.014}  & \\
    \arrayrulecolor{black!30}\cmidrule(lr){1-5}

    \multirow{7}{*}{\makecell{Building \\ Density}}
      & UNet\,/\,Micro UNet & 17.3\,/\,0.49 & \textbf{0.018}\,/\,0.024 & \textbf{0.006}\,/\,0.013  & \\
      & TerraMind-base (Full\,/\,Frozen\,/\,Embedding) & 102\,/\,15.5\,/\,0.39 & 0.023\,/\,0.024\,/\,0.025 & \textbf{0.009}\,/\,\textbf{0.009}\,/\,0.010  & \\
      & Prithvi v2-300 (Full\,/\,Frozen\,/\,Embedding) & 324\,/\,20.3\,/\,0.46 & \textbf{0.022}\,/\,0.024\,/\,0.029 & \textbf{0.008}\,/\,\textbf{0.009}\,/\,0.011 \\
      & DOFA-Large (Full\,/\,Frozen\,/\,Embedding)     & 357\,/\,20.3\,/\,0.46 & 0.024\,/\,0.026\,/\,0.023 & \textbf{0.009}\,/\,0.010\,/\,0.014 &  \\
      & LIANet-Base & 0.5 & \textbf{0.022} & \textbf{0.009}  & \\
      & LIANet-Large & 0.5 & \textbf{0.021} & \textbf{0.008}  & \\
    \bottomrule
    \end{tabular}
    }
    \label{tab:results_5k}
\end{table*}

\paragraph{Ablation: Scaling of the Encoded Area.}
\begin{figure}
    \centering
    \includegraphics[width=.70\linewidth]{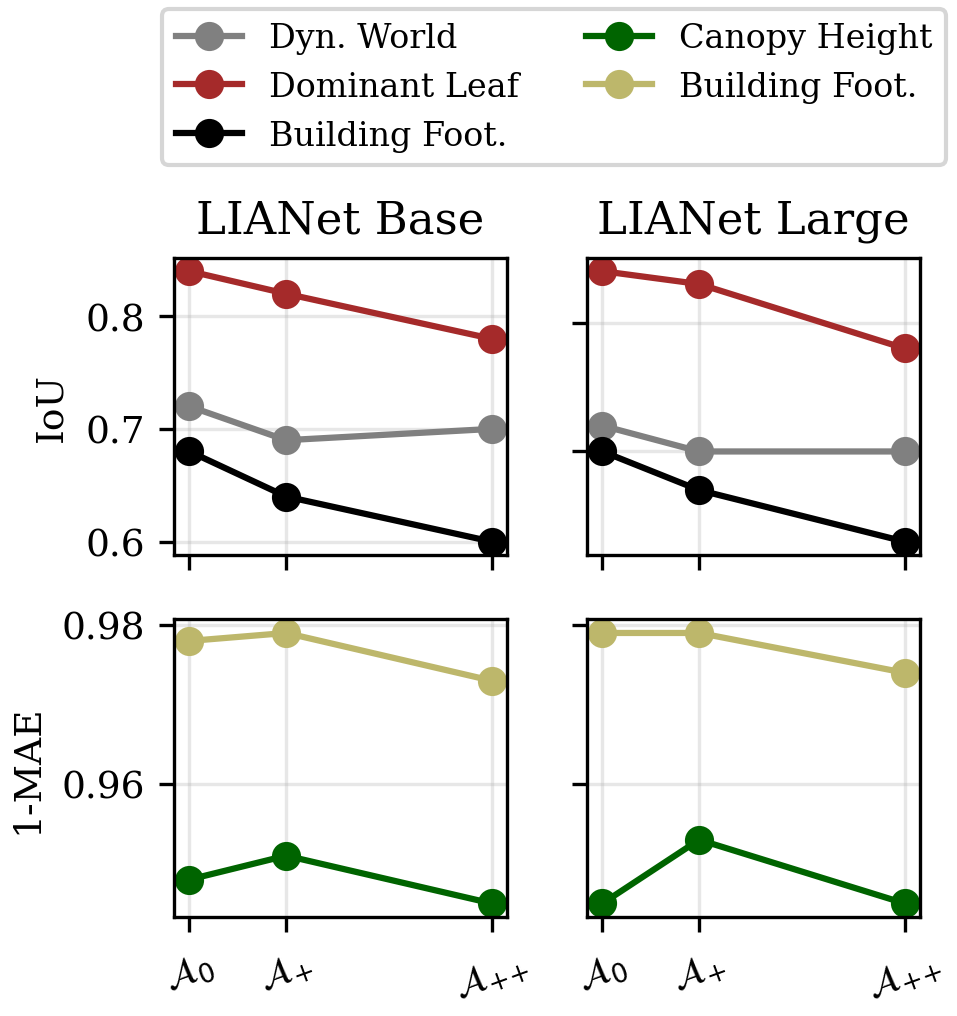}
    \caption{Performance across five downstream tasks as a function of the encoded area of interest $\mathcal{A}$ and as a function of the two model sizes. For regression tasks, we report $1-\text{MAE}$ to highlight the overall trend more clearly. Note that the regression task of building footprints reported in yellow is the prediction of the percentage of the buildings present in each pixel.}
    \label{fig:scaling}
\end{figure}
\Cref{fig:scaling} compares Intersection over Union (IoU) for pixel-wise classification tasks and Mean Absolute Error (MAE) for regression tasks of both \textit{LIANet} model sizes and all encoded areas. The results reveal a consistent trend of decreasing downstream performance with increasing area size, and a slight improvement in performance with larger model capacity, mirroring the reconstruction quality patterns observed in \cref{fig:reconstruction_visual}. The numerical results underlying \cref{fig:scaling} are provided in \cref{tab:results_5k} and in the \cref{tab:regression_7k} and \cref{tab:regression_10k} within \cref{sec:Suppmat_finetining_details} of the \textbf{supplementary materials}.

\section{Discussion}
\label{sec:discussion}

\paragraph{Downstream Task Performance.}
Across the evaluated downstream classification tasks, the proposed \textit{hyper-local} \textit{LIANet} is consistently competitive with the best baseline methods, despite the latter using significantly fewer parameters (see \cref{tab:results_5k}). To verify that these findings are not limited to our custom dataset, we further evaluate LIANet on two adapted PANGAEA benchmark datasets (see \cref{sec:pastis_burnscars} in the supplementary material). Under these independent benchmark protocols, LIANet maintains competitive performance compared to established \acp{GFM}. These observations align with recent \ac{EO} benchmarks, which demonstrate that training from scratch can remain competitive with large-scale pretrained \acp{GFM} (see \cite{jakubik2025terramind,marsocci2024pangaea}). For the regression task, \textit{LIANet} outperforms both the evaluated \acp{GFM} and Micro UNet. Overall, light-weight fine-tuning of \textit{LIANet} achieves performance comparable to full from-scratch training and surpasses the evaluated \ac{GFM} baselines, while eliminating the need for end users to access and preprocess data.

The primary limitation of LIANet lies in its spatial scope: it operates exclusively within the geographic region in which it was originally encoded and does not generalize to other areas. As discussed earlier, this restriction is only justified in scenarios where pretraining is conducted centrally, and the resulting region-specific models are distributed to a broad range of end users.

\paragraph{Model Size, Area Size, and Label Efficiency.} Two decoder configurations are evaluated: a base-sized decoder ($D_{base}$) and a larger variant ($D_{large}$), comprising $75\,$M and $133\,$M tunable parameters, respectively.
Although the larger decoder yields visibly improved reconstruction quality (see \cref{fig:reconstruction_visual}), its advantage is only marginal (compare \cref{tab:results_5k} and \cref{fig:scaling}). \Cref{fig:scaling} illustrates the effect of increasing the target area $\mathcal{A}$.
Both reconstruction quality (see \cref{fig:reconstruction_visual}) and downstream task performance (\cref{fig:scaling}, \cref{tab:results_5k}, and additional tables in \cref{sec:Suppmat_finetining_details} of supplementary material) exhibit a consistent trend of decreasing performance as the target area increases. As this study is a proof of concept, future research could explore more systematic scaling of model size and extend $\mathcal{A}$ from the municipality to the country level.

\paragraph{End-User Impact and Future Directions.}
This manuscript introduced the concept of \textit{hyper-local} continuous representations of \ac{EO} data. Similar to embedding-based workflows, our approach can be viewed as a provider-side abstraction that reduces the need for transferring and preprocessing large volumes of imagery. For end users, this enables adapting a pretrained regional model to downstream tasks without accessing raw data or aligning it with the requirements of large foundation models. While our initial results demonstrate strong performance, scaling and centralized pretraining remain important challenges to establish \textit{LIANet} as a practical alternative to existing approaches. Future work will focus on analyzing grid configurations, improving temporal representations for smoother transitions over time, and developing efficient update strategies for newly acquired data. We will also conduct a more comprehensive comparison with state-of-the-art embedding-based methods and location encoders.

\section{Conclusion}
\label{sec:conclusion}
We introduced \textit{LIANet}, a coordinate-based neural representation inspired by INRs to learn continuous spatiotemporal embeddings of the Earth’s surface. Pretrained generatively from individual $(x,y,t)$ coordinates, the model reconstructs multispectral imagery and enables dense few-shot adaptation to downstream tasks with \emph{minimal} tunable parameters. 
Designed as a provider-side abstraction, \textit{LIANet} allows end users to fine-tune models for their applications without accessing or preprocessing raw satellite data. Rather than pursuing global generalization, it serves a different audience: users who require a high-performing, region-specific model that eliminates data preparation overhead while preserving access to the underlying observations through reconstruction. The model remains lightweight, adaptable to diverse downstream tasks, and tailored to a given area and time of interest. Our extensive experiments demonstrate competitive performance across seven diverse applications. Future work will scale geographic coverage and temporal depth to further position \textit{LIANet} as a practical \ac{EO} system model.

\newpage

\noindent\textbf{Acknowledgments} \\

\noindent This research is partially funded by the Embed2Scale project, co-funded by the EU Horizon Europe programme (Grant Agreement No. 101131841), with support from the Swiss State Secretariat for Education, Research and Innovation (SERI) and UK Research and Innovation (UKRI).

{
    \small
    \bibliographystyle{ieeenat_fullname}
    \bibliography{main}
}

\clearpage
\setcounter{page}{1}
\maketitlesupplementary

\begin{table*}[ht!]
    \centering
    \footnotesize
    \caption{The pixel-wise classification performance evaluation of two datasets from the PANGAEA benchmark, measured with Intersection-over-Union (IoU), Accuracy (Acc), and F1-score (all calculated with macro averaging), along with the tunable parameter counts. For all metrics reported, the corresponding \textbf{top three} performances are printed \textbf{bold}.}
\resizebox{\textwidth}{!}{
    \begin{tabular}{clcccc}
    \toprule
    \textbf{Task} & \textbf{Model / Setting} & \textbf{\# Tunable Params (M)} & \textbf{IoU} & \textbf{ACC} & \textbf{F1} \\
    \midrule

    \multirow{5}{*}{\makecell{PASTIS}}
      & UNet\,/\,Micro UNet  & 17.3\,/\,0.49 & 0.18 \,/\,0.13 & 0.26\,/\,0.19 & 0.26\,/\,0.19 \\
      & TerraMind-base (Full\,/\,Frozen\,/\,Embedding) & 102\,/\,15.5\,/\,0.39 & \textbf{0.26}\,/\,0.23\,/\,\textbf{0.24} & \textbf{0.35}\,/\,0.30\,/\,\textbf{0.31} & \textbf{0.33}\,/\,\textbf{0.32}\,/\,\textbf{0.32} \\
      & Prithvi v2-300 (Full\,/\,Frozen\,/\,Embedding) & 324\,/\,20.3\,/\,0.46 & 0.23\,/\,0.21\,/\,0.19 & 0.29\,/\,0.28\,/\,0.25 & \textbf{0.32}\,/\,0.30\,/\,0.27 \\
      & DOFA-Large (Full\,/\,Frozen\,/\,Embedding)     & 357\,/\,20.3\,/\,0.46 & 0.19\,/\,0.17\,/\,0.14 & 0.23\,/\,0.24\,/\,0.19 & 0.28\,/\,0.25\,/\,0.19 \\
      & LIANet-Modified & 0.87 & \textbf{0.34} & \textbf{0.46} & \textbf{0.46} \\
    \arrayrulecolor{black!30}\cmidrule(lr){1-6}
    \multirow{5}{*}{\makecell{Burn Scars}}
      & UNet\,/\,Micro UNet & 17.3\,/\,0.49 & \textbf{0.45}\,/\,\textbf{0.47}& \textbf{0.65}\,/\,\textbf{0.62} & \textbf{0.62}\,/\,\textbf{0.62}\\
      & TerraMind-base (Full\,/\,Frozen\,/\,Embedding) & 92\,/\,5.0\,/\,0.39 & 0.39\,/\,0.41\,/\,0.35 & 0.58\,/\,0.60\,/\,0.52 & 0.55\,/\,0.57\,/\,0.51 \\
      & Prithvi v2-300 (Full\,/\,Frozen\,/\,Embedding) & 313\,/\,7.7\,/\,0.46 & \textbf{0.44}\,/\,0.43\,/\,0.42 & 0.60\,/\,\textbf{0.63}\,/\,0.61 & \textbf{0.59}\,/\,\textbf{0.59}\,/\,0.58 \\
      & DOFA-Large (Full\,/\,Frozen\,/\,Embedding)     & 357\,/\,7.7\,/\,0.46 & 0.37\,/\,0.41\,/\,0.38 & 0.59\,/\,0.59\,/\,0.55 & 0.51\,/\,0.58\,/\,0.55 \\
      & LIANet-Modified & 0.87 & \textbf{0.47} & \textbf{0.63} & \textbf{0.63} \\
    \bottomrule
    \end{tabular}
    }
    \label{tab:pangeae-benchmark-results}
\end{table*}

\begin{table*}[ht!]
    \centering
    \footnotesize
    \caption{(Top) Pixel-wise classification task performance measured with Intersection-over-Union (IoU), Accuracy (Acc), and F1-score (all calculated with macro averaging) along with the tunable parameter counts for the area of interest $\mathcal{A}_{+}$.
    (Bottom) Regression task performance (Mean Absolute Error (MAE) and Mean Squared Error (MSE)) along with the tunable parameter counts for the area of interest $\mathcal{A}_{+}$. The from-scratch trainings and \acp{FM} are evaluated in different configurations with a varying number of tunable parameters.
    For all metrics reported, the corresponding \textbf{top three} performances are printed \textbf{bold}. Due to the limited number of comparisons, we are not marking any models for the task of building footprint segmentation.}
\resizebox{\textwidth}{!}{
    \begin{tabular}{clcccc}
    \toprule
    \textbf{Task} & \textbf{Model / Setting} & \textbf{\# Tunable Params (M)} & \textbf{IoU} & \textbf{ACC} & \textbf{F1} \\
    \midrule

    \multirow{6}{*}{\makecell{Dynamic \\ World}}
      & UNet\,/\,Micro UNet  & 17.3\,/\,0.49 & \textbf{0.75}\,/\,0.66 & \textbf{0.82}\,/\,0.76 & \textbf{0.84}\,/\,0.76 \\
      & TerraMind-base (Full\,/\,Frozen\,/\,Embedding) & 102\,/\,15.5\,/\,0.39 & \textbf{0.70}\,/\,\textbf{0.70}\,/\,0.65 & \textbf{0.78}\,/\,0.77\,/\,0.72 & \textbf{0.80}\,/\,\textbf{0.80}\,/\,0.72 \\
      & Prithvi v2-300 (Full\,/\,Frozen\,/\,Embedding) & 324\,/\,20.3\,/\,0.46 & \textbf{0.69}\,/\,0.65\,/\,0.58 & 0.77\,/\,0.73\,/\,0.68 & \textbf{0.79}\,/\,0.75\,/\,0.67 \\
      & DOFA-Large (Full\,/\,Frozen\,/\,Embedding)     & 357\,/\,20.3\,/\,0.46 & 0.68\,/\,0.62\,/\,0.46 & 0.75\,/\,0.71\,/\,0.54 & 0.78\,/\,0.72\,/\,0.56 \\
      & LIANet-Base & 0.5 & \textbf{0.69} & \textbf{0.80} & \textbf{0.79} \\
      & LIANet-Large & 0.5 & \textbf{0.70} & \textbf{0.80} & \textbf{0.80} \\
    \arrayrulecolor{black!30}\cmidrule(lr){1-6}

    \multirow{6}{*}{\makecell{Dominant \\ Leaf Type}}
      & UNet\,/\,Micro UNet  & 17.3\,/\,0.49 & \textbf{0.82}\,/\,0.78 & \textbf{0.89}\,/\,0.86 & \textbf{0.90}\,/\,\textbf{0.87} \\
      & TerraMind-base (Full\,/\,Frozen\,/\,Embedding) & 102\,/\,15.5\,/\,0.39 & \textbf{0.79}\,/\,0.78\,/\,0.75 & \textbf{0.87}\,/\,0.85\,/\,0.84 & \textbf{0.88}\,/\,0.86\,/\,0.85 \\
      & Prithvi v2-300 (Full\,/\,Frozen\,/\,Embedding) & 324\,/\,20.3\,/\,0.46 & \textbf{0.79}\,/\,0.76\,/\,0.70 & \textbf{0.87}\,/\,0.84\,/\,0.79 & \textbf{0.88}\,/\,0.85\,/\,0.80 \\
      & DOFA-Large (Full\,/\,Frozen\,/\,Embedding)     & 357\,/\,20.3\,/\,0.46 & 0.76\,/\,0.71\,/\,0.62 & 0.85\,/\,0.80\,/\,0.72 & 0.86\,/\,0.81\,/\,0.73 \\
      & LIANet-Base & 0.5 & \textbf{0.82} & \textbf{0.90} & \textbf{0.90} \\
      & LIANet-Large & 0.5 & \textbf{0.83} & \textbf{0.90} & \textbf{0.90} \\
    \arrayrulecolor{black!30}\cmidrule(lr){1-6}

    \multirow{3}{*}{\makecell{Building Footprint \\ Segmentation}}
      & UNet\,/\,Micro UNet  & 17.3\,/\,0.49 & 0.76\,/\,0.68 & 0.87\,/\,0.79 & 0.84\,/\,0.77 \\
      & LIANet-Base & 0.5 & 0.64 & 0.71 & 0.72 \\
      & LIANet-Large & 0.5 & 0.67 & 0.87 & 0.76 \\

    \arrayrulecolor{black!100}\toprule
    \textbf{Task} & \textbf{Model / Setting} & \textbf{\# Tunable Params (M)} & \textbf{MAE} & \textbf{MSE} &  \\
    \midrule

    \multirow{6}{*}{\makecell{Canopy \\ Height}}
      & UNet\,/\,Micro UNet  & 17.3\,/\,0.49 & \textbf{0.049}\,/\,0.065 & \textbf{0.012}\,/\,0.019 \\
      & TerraMind-base (Full\,/\,Frozen\,/\,Embedding) & 102\,/\,15.5\,/\,0.39 & \textbf{0.048}\,/\,0.050\,/\,0.110 & \textbf{0.012}\,/\,\textbf{0.012}\,/\,0.056 \\
      & Prithvi v2-300 (Full\,/\,Frozen\,/\,Embedding) & 324\,/\,20.3\,/\,0.46 & \textbf{0.048}\,/\,0.053\,/\,0.110 & \textbf{0.012}\,/\,0.014\,/\,0.056 \\
      & DOFA-Large (Full\,/\,Frozen\,/\,Embedding)     & 357\,/\,20.3\,/\,0.46 & 0.051\,/\,0.057\,/\,0.110 & \textbf{0.013}\,/\,0.016\,/\,0.056 \\

      & LIANet-Base & 0.5 & \textbf{0.049} & \textbf{0.011} \\
      & LIANet-Large & 0.5 & \textbf{0.047} & \textbf{0.011} \\
    \arrayrulecolor{black!30}\cmidrule(lr){1-5}

    \multirow{6}{*}{\makecell{Building \\ Density}}
      & UNet\,/\,Micro UNet  & 17.3\,/\,0.49 & \textbf{0.017}\,/\,0.022 & \textbf{0.006}\,/\,0.012 \\
      & TerraMind-base (Full\,/\,Frozen\,/\,Embedding) & 102\,/\,15.5\,/\,0.39 & 0.022\,/\,0.023\,/\,0.023 & \textbf{0.008}\,/\,\textbf{0.009}\,/\,\textbf{0.009} \\
      & Prithvi v2-300 (Full\,/\,Frozen\,/\,Embedding) & 324\,/\,20.3\,/\,0.46 & \textbf{0.020}\,/\,0.023\,/\,0.027 & \textbf{0.008}\,/\,\textbf{0.008}\,/\,0.010 \\
      & DOFA-Large (Full\,/\,Frozen\,/\,Embedding)     & 357\,/\,20.3\,/\,0.46 & 0.022\,/\,0.024\,/\,\textbf{0.021} & \textbf{0.008}\,/\,\textbf{0.009}\,/\,0.013 \\
      & LIANet-Base & 0.5 & \textbf{0.021} & \textbf{0.009} \\
      & LIANet-Large & 0.5 & \textbf{0.021} & \textbf{0.008} \\
    \bottomrule
    \end{tabular}
    }
    \label{tab:regression_7k}
\end{table*}

In this supplementary material, we provide additional technical and experimental details that complement the main paper. 
\begin{itemize}
    \item In \cref{sec:pretraining_steup_hashcollision}, we describe the pretraining configuration of \textit{LIANet}, including optimization settings and an analysis of hash collisions in the multiresolution hash tables. 
    \item In \cref{sec:Suppmat_finetining_details}, we outline the fine-tuning strategy across the downstream \ac{EO} tasks presented in the main paper, including dataset descriptions, evaluation protocols, and complete results tables for $\mathcal{A}_{+}$ and $\mathcal{A}_{++}$. We further provide qualitative predictions and an ablation study on the effect of varying the fine-tuning area size.
    \item In \cref{sec:pastis_burnscars}, we report fine-tuning results on two datasets from the \textit{PANGAEA} benchmark~\cite{marsocci2024pangaea}, and detail the corresponding experimental setup.
    \item Finally, in \cref{sec:neural_compression}, we analyze \textit{LIANet} from a neural compression perspective, quantifying its implicit encoding efficiency and reconstruction fidelity.
\end{itemize}
\textbf{The source code and the dataset that is used for pretraining and fine-tuning of the proposed framework can be found in \url{https://github.com/mojganmadadi/LIANet/tree/v1.0.1}.}
\section{Pretraining Setup}\label{sec:pretraining_steup_hashcollision}
To pretrain the proposed \textit{LIANet} on different-sized areas, we use $L_1$ loss, AdamW optimizer with a base learning rate of $5~\times~10^{-4}$, and a Cosine learning rate scheduler that has 5 warm-up epochs. Every epoch is backpropagated with 1,024,000 randomly selected points, trained with a batch size of 64. The number of training epochs for $\mathcal{A}_{0}$ is 225, for $\mathcal{A}_{+}$ is 393, and for $\mathcal{A}_{++}$ is 897 where the training converges. 
The encoder layers of \textit{LIANet-Base} contain \textit{ResNet50} blocks, whereas \textit{LIANet-Large} advantages from deeper layers of \textit{ResNet101}.

\paragraph{Hash Collisions.}
Monitoring hash collisions is essential to ensure that the encoded embedding $\mathbf{g}_{x,y,t}$ remains discriminative for each spatial coordinate $(x,y)$. 
At resolution level $\ell \in \{1,\dots,L\}$ with hash function $h_\ell(\cdot)$, a collision occurs when two distinct grid nodes map to the same table index:
\begin{equation}
(x_i, y_i) \neq (x_j, y_j) 
\quad \text{and} \quad 
h_\ell(x_i, y_i) = h_\ell(x_j, y_j).
\end{equation}
Collisions become more likely at higher resolutions, where the number of grid nodes exceeds the hash table length. In principle, this may lead to conflicting gradient updates if spatially distant locations alias across all resolution levels and therefore share identical concatenated embeddings.
However, the final representation is constructed by concatenating multi-resolution encodings, each generated with an independent random seed (see \texttt{LIANet/Pretraining/src/models/LIANet.py}). Consequently, simultaneous collisions across all levels are statistically unlikely. The chosen hyperparameters balance memory efficiency and representational capacity while keeping the empirical collision rate negligible.

\section{Fine-Tuning on \texorpdfstring{$\mathcal{A}_{+}$}{A+} and \texorpdfstring{$\mathcal{A}_{++}$}{A++}}
\label{sec:Suppmat_finetining_details}

\begin{figure*}
    \centering
    \includegraphics[width=\linewidth]{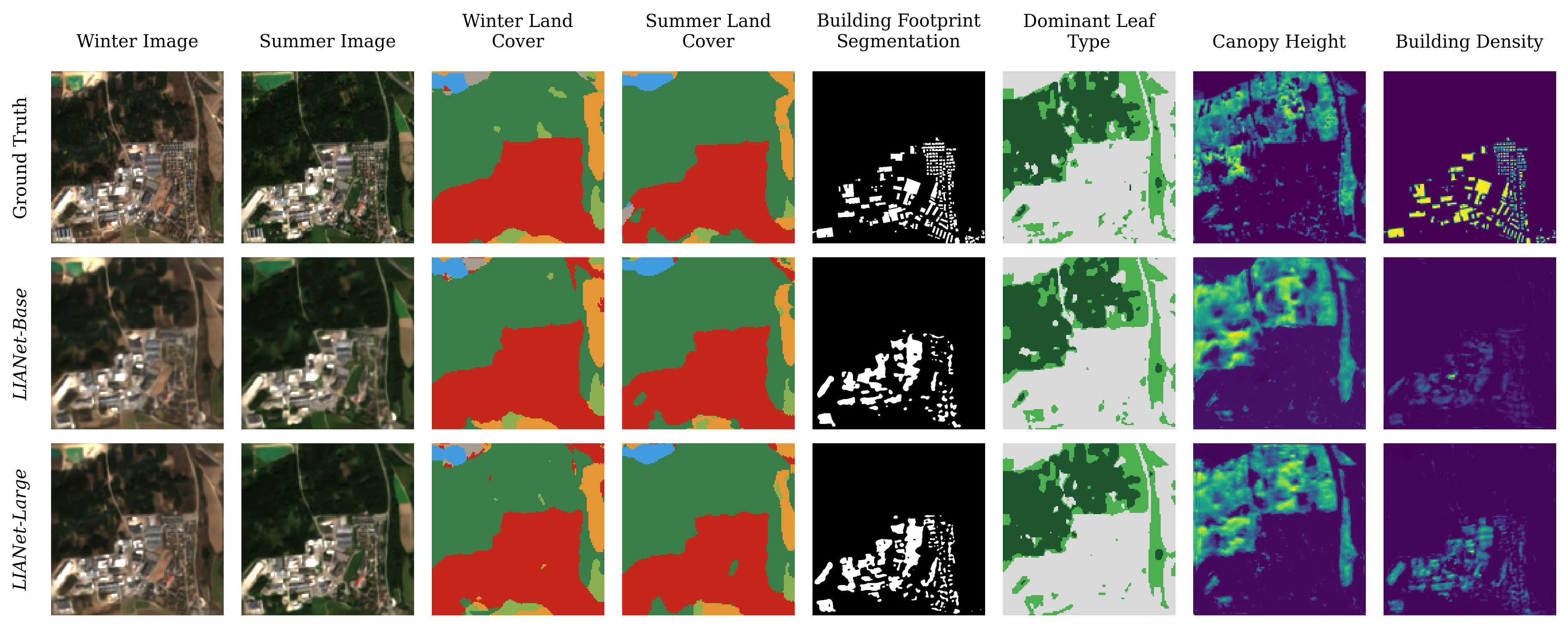}
    \includegraphics[width=\linewidth]{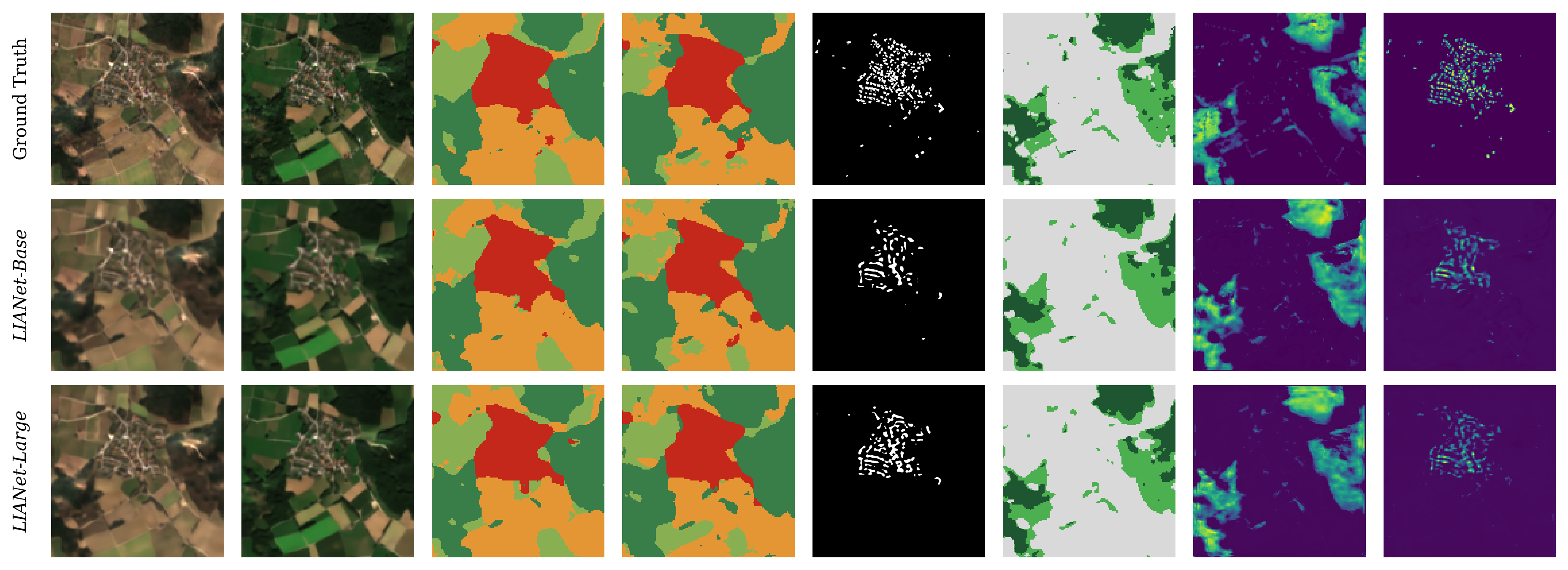}
    \includegraphics[width=\linewidth]{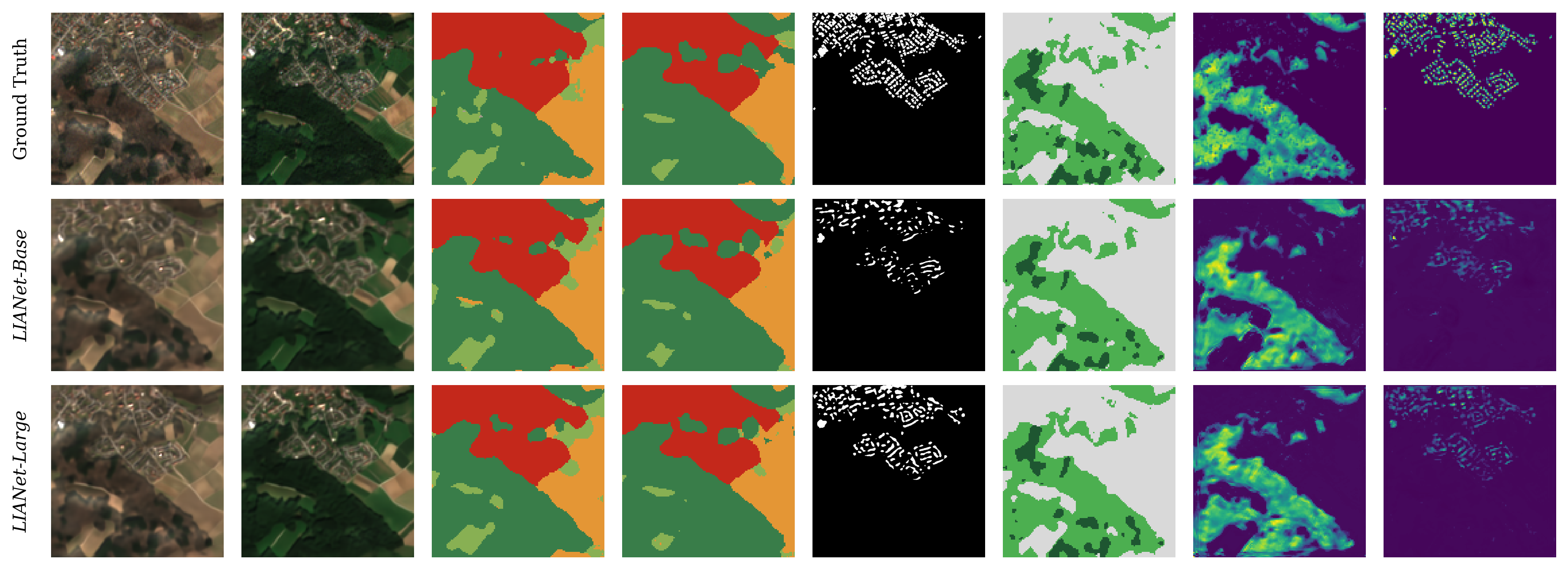}
    \includegraphics[width=\linewidth]{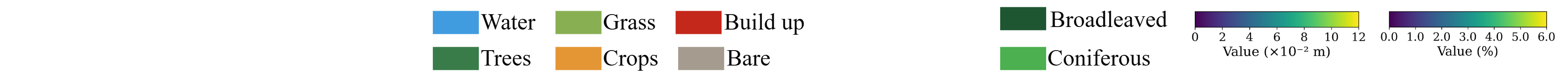}
    \caption{Three example visualization of \textit{LIANet-Base} and \textit{LIANet-Large} reconstruction and prediction performances on downstream applications. The landcover labels have seasonal predictions, whereas the other tasks have a single label for all timestamps. The improvement in the prediction of \textit{LIANet-Large} can clearly be seen in the tasks concerning the segmentation and density estimation of building footprints.}
    \label{fig:visual_results_of_dsa}
\end{figure*}

We evaluate on five downstream tasks, including regression, binary, and multi-class segmentation, to assess the utility of the learned representations. Three visual sample patches from two seasons together with their reconstruction results of \textit{LIANet-Base} and \textit{LIANet-Large}, as well as their predicted labels for all tasks, are illustrated in \cref{fig:visual_results_of_dsa}.

\paragraph{Dynamic World Land Cover.}
Dynamic World provides near-real-time, \SI{10}{m} global land-cover maps predicted from Sentinel-2 imagery \cite{brown2022dynamic}. Each scene is labeled into nine classes (Water, Trees, Grass, Crops, Shrub \& Scrub, Flooded Vegetation, Built-up, Bare Ground, Snow \& Ice). For our area of interest, we form a 6-class pixel-wise segmentation (0: Water, 1: Trees, 2: Grass, merging Shrub \& Scrub, 3: Crops, 4: Built-up, 5: Bare—merging Snow \& Ice). Because labels are available per timestamp, this task probes how temporal embeddings are exploited during fine-tuning.

\paragraph{Canopy Height.}
We use canopy height maps derived from high-resolution Maxar imagery \cite{meta_wri_chm_2024}. This is a single-output regression task with one label per location across all timestamps, evaluating absolute height estimation from multispectral inputs.

\paragraph{Dominant Leaf Type.}
We use the Copernicus High Resolution Layer on tree cover/forests \cite{clms_dlt_2018_present} labels that include three classes: broadleaf, coniferous, and no-forest. This multi-class segmentation leverages all 12 Sentinel-2 bands and tests the model’s ability to discriminate forest functional types.

\paragraph{Building Coverage Percentage.}
We estimate the fraction of building coverage per \SI{10}{m}\,$\times$\,\SI{10}{m} pixel using labels derived from Microsoft building footprints (from Maxar/Airbus) \cite{MBF}. This is a scalar regression task that requires resolving sub-pixel structure from a multispectral context.

\paragraph{Building Footprint Segmentation.}
We perform binary building segmentation at \SI{2.5}{m} ground sampling. This task evaluates transfer to higher spatial resolution quality beyond the \SI{10}{m} Sentinel-2 scale.

\textbf{Fine-tuning Setup.}
We fine-tune both \textit{LIANet-Base} and \textit{LIANet-Large} across all downstream tasks using a batch size of~32 and input patches of \(128\times128\) pixels. Each task is trained for 50~epochs with a Cosine learning rate scheduler that has 5 warm-up epochs. We employ cross-entropy loss for segmentation tasks, L1~loss for canopy height regression, and Huber~loss for building coverage estimation. Learning rates are set to \(1\times10^{-3}\) for segmentation and \(5\times10^{-5}\) for regression. The training regions correspond to \SI{500}{km^2} within each study area, resulting in 20\% of~\(\mathcal{A}_{0}\), 10\% of~\(\mathcal{A}_{+}\), and 4\% of~\(\mathcal{A}_{++}\) used for training, with the remainder reserved for validation. This limited-data setup is designed to evaluate the robustness and data efficiency of the learned representations.

All benchmark models are trained under consistent configurations: batch size of 32, input patch size of \(128\times128\) pixels, and 50~epochs. We employ cross-entropy loss for segmentation tasks, L1 loss for canopy height regression, and Huber loss for building density estimation. Optimization is performed using AdamW with a learning rate of \(1\times10^{-4}\) for segmentation and \(5\times10^{-5}\) for regression tasks. Training and validation splits follow the same setup as defined in \cref{sec:Suppmat_finetining_details}.
\begin{table}[h!]
    \centering
    \footnotesize
    \caption{Comparison of the pixel-wise land cover classification performance measured with Intersection-over-Union (IoU), Accuracy (Acc), and F1-score (all calculated with macro averaging) with respect to the different training area sizes.}
    \begin{tabular}{lcccc}
        \toprule
        \textbf{Model} & \textbf{Train Area (\SI{}{\square\kilo\metre})} & \textbf{IoU} & \textbf{ACC} & \textbf{F1} \\
        \midrule
        \multirow{4}{*}{LIANet-Base} 
            & 80 & 0.62 & 0.70 & 0.72 \\
            & 160 & 0.66 & 0.75 & 0.75 \\
            & 320 & 0.71 & 0.79 & 0.80 \\
            & 500 & 0.72 & 0.82 & 0.81 \\
            & 640 & 0.73 & 0.82 & 0.82 \\
        \arrayrulecolor{black!30}\cmidrule(lr){1-5}
        \multirow{4}{*}{LIANet-Large} 
            & 80 & 0.63 & 0.71 & 0.73 \\
            & 160 & 0.65 & 0.74 & 0.75 \\
            & 320 & 0.71 & 0.79 & 0.80 \\
            & 500 & 0.72 & 0.81 & 0.81 \\
            & 640 & 0.73 & 0.82 & 0.82 \\
        \bottomrule
    \end{tabular}
    \label{tab:dynamic_world_samples}
\end{table}

\begin{table*}[t!]
    \centering
    \footnotesize
        \caption{(Top) Pixel-wise classification task performance measured with Intersection-over-Union (IoU), Accuracy (Acc), and F1-score (all calculated with macro averaging) along with the tunable parameter counts for the area of interest $\mathcal{A}_{++}$.
    (Bottom) Regression task performance (Mean Absolute Error (MAE) and Mean Squared Error (MSE)) along with the tunable parameter counts for the area of interest $\mathcal{A}_{++}$. The from-scratch trainings and \acp{FM} are evaluated in different configurations with a varying number of tunable parameters.
    For all metrics reported, the corresponding \textbf{top three} performances are printed \textbf{bold}. Due to the limited number of comparisons, we are not marking any models for the task of building footprint segmentation.}
\resizebox{\textwidth}{!}{
    \begin{tabular}{clcccc}
    \toprule
    \textbf{Task} & \textbf{Model / Setting} & \textbf{\# Tunable Params (M)} & \textbf{IoU} & \textbf{ACC} & \textbf{F1} \\
    \midrule

    \multirow{6}{*}{\makecell{Dynamic \\ World}}
      & UNet\,/\,Micro UNet  & 17.3\,/\,0.49 & \textbf{0.76}\,/\,0.68 & \textbf{0.83}\,/\,0.76 & \textbf{0.85}\,/\,0.78 \\
      & TerraMind-base (Full\,/\,Frozen\,/\,Embedding) & 102\,/\,15.5\,/\,0.39 & \textbf{0.73}\,/\,\textbf{0.72}\,/\,0.63 & \textbf{0.79}\,/\,\textbf{0.78}\,/\,0.70 & \textbf{0.82}\,/\,\textbf{0.81}\,/\,0.71 \\
      & Prithvi v2-300 (Full\,/\,Frozen\,/\,Embedding) & 324\,/\,20.3\,/\,0.46 & 0.71\,/\,0.68\,/\,0.61 & \textbf{0.78}\,/\,0.75\,/\,0.70 & 0.80\,/\,0.77\,/\,0.70 \\
      & DOFA-Large (Full\,/\,Frozen\,/\,Embedding)     & 357\,/\,20.3\,/\,0.46 & 0.67\,/\,0.64\,/\,0.42 & 0.73\,/\,0.72\,/\,0.51 & 0.77\,/\,0.74\,/\,0.53 \\
      & LIANet-Base & 0.5 & 0.70 & \textbf{0.78} & 0.79 \\
      & LIANet-Large & 0.5 & 0.70 & \textbf{0.78} & 0.80 \\
    \arrayrulecolor{black!30}\cmidrule(lr){1-6}

    \multirow{6}{*}{\makecell{Dominant \\ Leaf Type}}
      & UNet\,/\,Micro UNet  & 17.3\,/\,0.49 & \textbf{0.79}\,/\,0.76 & \textbf{0.88}\,/\,0.85 & \textbf{0.88}\,/\,0.85 \\
      & TerraMind-base (Full\,/\,Frozen\,/\,Embedding) & 102\,/\,15.5\,/\,0.39 & \textbf{0.77}\,/\,0.75\,/\,0.73 & \textbf{0.86}\,/\,0.83\,/\,0.82 & \textbf{0.86}\,/\,0.85\,/\,0.83 \\
      & Prithvi v2-300 (Full\,/\,Frozen\,/\,Embedding) & 324\,/\,20.3\,/\,0.46 & \textbf{0.77}\,/\,0.73\,/\,0.68 & \textbf{0.86}\,/\,0.82\,/\,0.78 & \textbf{0.86}\,/\,0.83\,/\,0.79 \\
      & DOFA-Large (Full\,/\,Frozen\,/\,Embedding)     & 357\,/\,20.3\,/\,0.46 & 0.74\,/\,0.69\,/\,0.59 & 0.83\,/\,0.78\,/\,0.70 & 0.84\,/\,0.79\,/\,0.71 \\
      & LIANet-Base & 0.5 & \textbf{0.78} & \textbf{0.87} & \textbf{0.87} \\
      & LIANet-Large & 0.5 & \textbf{0.78} & \textbf{0.87} & \textbf{0.87} \\
    \arrayrulecolor{black!30}\cmidrule(lr){1-6}

    \multirow{3}{*}{\makecell{Building Footprint \\ Segmentation}}
      & UNet\,/\,Micro UNet  & 17.3\,/\,0.49 & 0.75\,/\,0.67 & 0.88\,/\,0.79 & 0.84\,/\,0.76 \\
      & LIANet-Base & 0.5 & 0.60 & 0.65 & 0.68 \\
      & LIANet-Large & 0.5 & 0.63 & 0.83 & 0.72 \\

    \arrayrulecolor{black!100}\toprule
    \textbf{Task} & \textbf{Model / Setting} & \textbf{\# Tunable Params (M)} & \textbf{MAE} & \textbf{MSE} &  \\
    \midrule

    \multirow{6}{*}{\makecell{Canopy \\ Height}}
      & UNet\,/\,Micro UNet  & 17.3\,/\,0.49 & \textbf{0.053}\,/\,0.069 & \textbf{0.013}\,/\,0.020 \\
      & TerraMind-base (Full\,/\,Frozen\,/\,Embedding) & 102\,/\,15.5\,/\,0.39 & \textbf{0.052}\,/\,0.056\,/\,0.120 & \textbf{0.013}\,/\,\textbf{0.014}\,/\,0.060 \\
      & Prithvi v2-300 (Full\,/\,Frozen\,/\,Embedding) & 324\,/\,20.3\,/\,0.46 & \textbf{0.055}\,/\,0.058\,/\,0.120 & \textbf{0.014}\,/\,\textbf{0.015}\,/\,0.060 \\
      & DOFA-Large (Full\,/\,Frozen\,/\,Embedding)     & 357\,/\,20.3\,/\,0.46 & 0.058\,/\,0.062\,/\,0.120 & \textbf{0.015}\,/\,0.017\,/\,0.060 \\
      & LIANet-Base & 0.5 & \textbf{0.055} & \textbf{0.013} \\
      & LIANet-Large & 0.5 & \textbf{0.055} & \textbf{0.013} \\
    \arrayrulecolor{black!30}\cmidrule(lr){1-5}

    \multirow{6}{*}{\makecell{Building \\ Density}}
      & UNet\,/\,Micro UNet  & 17.3\,/\,0.49 & \textbf{0.021}\,/\,0.027 & \textbf{0.008}\,/\,0.016 \\
      & TerraMind-base (Full\,/\,Frozen\,/\,Embedding) & 102\,/\,15.5\,/\,0.39 & 0.027\,/\,0.028\,/\,0.028 & \textbf{0.011}\,/\,\textbf{0.011}\,/\,\textbf{0.011} \\
      & Prithvi v2-300 (Full\,/\,Frozen\,/\,Embedding) & 324\,/\,20.3\,/\,0.46 & \textbf{0.025}\,/\,0.028\,/\,0.033 & \textbf{0.010}\,/\,\textbf{0.011}\,/\,0.012 \\
      & DOFA-Large (Full\,/\,Frozen\,/\,Embedding)     & 357\,/\,20.3\,/\,0.46 & 0.027\,/\,0.029\,/\,\textbf{0.025} & \textbf{0.011}\,/\,0.012\,/\,0.016 \\
      & LIANet-Base & 0.5 & 0.027 & 0.012 \\
      & LIANet-Large & 0.5 & \textbf{0.026} & 0.012 \\
    \bottomrule
    \end{tabular}
    }
    \label{tab:regression_10k}
\end{table*}

\paragraph{Ablation: Varying amount of annotated data during Fine-tuning.}
\Cref{tab:dynamic_world_samples} presents the performance of the \textit{LIANet-Base} and \textit{LIANet-Large} models when fine-tuned on varying fractions of the target area for pixel-wise land cover classification. This experiment investigates how much the training area can be reduced while still maintaining reasonable accuracy.  
As shown in \cref{tab:dynamic_world_samples}, although performance decreases with smaller training regions, both models still achieve satisfactory results even when fine-tuned on only $3\%$ of the target area,~$\mathcal{A}_{0}$.

\cref{tab:dynamic_world_samples} shows the performance of the \textit{LIANet-Base} and \textit{LIANet-Large} models in few-shot settings where the fine-tuning area varies from 25\% to 3\% of the target region. As observed, performance decreases as the fine-tuning area becomes smaller, yet both models maintain reasonable accuracy even with limited training data. This property is particularly valuable in scenarios with scarce labeled samples.

\section{Fine-Tuning on Standard Benchmark Datasets}
\label{sec:pastis_burnscars}
Selecting appropriate benchmarks for evaluating \textit{LIANet} requires compatibility between the benchmark input modality and the pretraining modality of \textit{LIANet}, the availability of georeferencing, and extensive spatial coverage. Since our framework models multispectral Sentinel-2 data within geographically contiguous regions, benchmarks must provide compatible imagery and sufficient spatial density.
Datasets such as \textit{Five Billion Pixels}~\cite{fivebillionpixels_FBP2023}, \textit{DynamicEarthNet}~\cite{dynamicearthnet_Toker_2022_CVPR}, and \textit{SpaceNet~7}~\cite{spacenet_7_van2021multi} are based on very high-resolution commercial imagery (e.g., Gaofen-2, PlanetFusion, Planet), which differs substantially from the Sentinel-2 modality considered in this work. In addition, some of these datasets are not fully open-access, limiting their suitability for large-scale generative pretraining.
Other benchmarks, such as \textit{MADOS}~\cite{mados_kikaki2024detecting}, predominantly contain low-frequency spatial content (e.g., water bodies), making them less suitable for assessing the high-frequency reconstruction capabilities of \textit{LIANet}. Furthermore, several datasets included in \textit{GeoBench}~\cite{lacoste2023geo} lack explicit georeferencing or require extensive preprocessing to ensure spatial alignment, which conflicts with the coordinate-based design of our approach.
We therefore select two datasets from the \textit{PANGAEA} benchmark~\cite{marsocci2024pangaea}, namely \textit{PASTIS}~\cite{pastis_garnot2021panoptic} and \textit{HLS BurnScars}~\cite{HLS_Foundation_2023}, and adapt them to match the experimental protocol of this study.

\paragraph{\textit{LIANet-Modified}.}
For benchmarking on \textit{PASTIS} and \textit{HLS BurnScars}, we introduce a modified configuration of \textit{LIANet}. Since pretraining spans multiple locations and a larger number of timestamps, we increase the encoding capacity while keeping the overall parameter count comparable to \textit{LIANet-Base}. The modified variant, referred to as \textit{LIANet-Modified}, uses a feature dimension of $F = 128$, a hash table size of $T = 2^{19}$, and $N_{\text{grids}} = 13$ resolution levels covering a complete Sentinel-2 tile. As in the original setup, 12 spectral channels are reconstructed. To maintain a comparable total parameter count, the CNN decoder head is reduced accordingly.

\paragraph{\textit{PASTIS} dataset} provides 19-class panoptic annotations of agricultural parcels together with multi-temporal Sentinel-2 image patches derived from four Sentinel-2 tiles in France. In total, the dataset contains 2,433 labeled patches.
For our experiments, we select two of the four available tiles, namely \texttt{T31TFM} and \texttt{T32ULU}, and use \emph{all} available patches and timestamps within these tiles. This corresponds to 1,279 patches, i.e., 52.6\% of the full dataset. Within the selected tiles, 19 cloud-free timestamps are available for one location and 18 for the other. \textit{LIANet-Modified} is pretrained on all timestamps of both tiles.
The labels are provided at 10\,m GSD with patch size $128 \times 128$ and are partitioned into five folds for cross-validation. We follow the official fold protocol, selecting each fold once for validation. The results reported in \cref{tab:pangeae-benchmark-results} correspond to the average performance across both tiles and all validation folds.

\paragraph{\textit{HLS Burn Scars} dataset} contains 804 labeled wildfire burn-scar patches derived from Harmonized Landsat and Sentinel-2 (HLS) imagery (2018–2021), sparsely distributed across the United States.
Since \textit{LIANet} models a geographically contiguous Sentinel-2 tile, we select the tile containing the largest number of labeled samples. Within this tile, seven training patches and two validation patches are available, resulting in a total of 9 labeled patches used for fine-tuning. This corresponds to approximately 1.1\% of the full labeled dataset.
For each labeled patch, a cloud-free post-event Sentinel-2 acquisition is used. Temporally close labels may share the same post-event image. In total, seven timestamps are used for generative pretraining over the selected tile.
The original train/validation split is preserved. To ensure spatial consistency, label masks are aligned to the generated imagery by matching resolution, coordinate reference system, and patch dimensions. The resulting performance is reported in \cref{tab:pangeae-benchmark-results}.
\paragraph{Pretraining and Fine-Tuning Setup.} For the pretraining on both tasks, the loss, optimizer, scheduler, and batch size are as reported in \cref{sec:pretraining_steup_hashcollision}. We implement the fine-tuning on 50 epochs with cross-entropy loss, AdamW optimizer, batch size of 32, and a fixed learning rate of \(1\times10^{-4}\). Baselines are fine-tuned using the same setup as \textit{LIANet}, similar to \cref{sec:Suppmat_finetining_details}. To improve learning stability for the end-to-end \ac{GFM} methods on the relatively small BurnScars dataset, we reduce the number of channels in the U-Net decoder. This decreases the number of trainable parameters compared to the other benchmark tasks.
\newpage
\section{Neural Compression Analysis}\label{sec:neural_compression}
Implicit neural representations are related to the field of neural compression \cite{dupont2021coin,gomes2025lossy}. Although \textit{LIANet} is not meant to serve as an \ac{EO} image compressor, we study compression performance in terms of reconstruction error as complementary information. Given coordinates $(x,y,t)$, \textit{LIANet} is able to regenerate images $\mathbf{I}$ over an area $\mathcal{A}$. Consequently, the image data is encoded by $D\circ E$ in a number of neural network parameters consuming a given amount of $B$ bytes of disk space. Provided the number of image channels $C$(=12 for Sentinel-2 L2A), and assuming a total number of $P_c$ spatial pixels in channel $c=1\dots C$ over area $\mathcal{A}$ for which $Q$ number of timestamps have been encoded, we derive the bits per pixel (bpp) according to:
\begin{equation}\label{eq:bpp}
1/\text{bpp} = \frac{Q}{8B} \sum_{c=1}^CP_c
\quad.
\end{equation}
It is worth noting that bpp$~\propto1/Q$. Compression methods that efficiently handle the temporal dimension $t$ in $(x,y,t)$ have the potential to significantly lower the bpp. \textit{LIANet}'s number of parameters to encode time, $Q\cdot N_\text{grid}\cdot F$, is substantially smaller compared to the remaining number of parameters of the network, in particular wrt.\ the decoder $D$.
\begin{figure}[h]
    \centering
    \includegraphics[width=\linewidth]{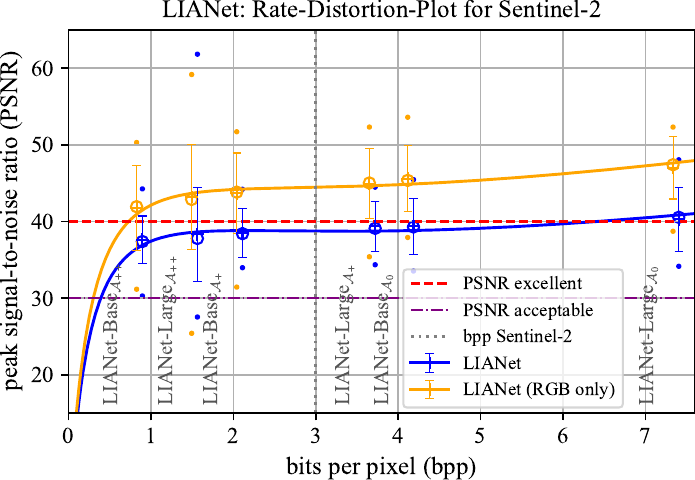}
    \caption{\textit{Rate-Distortion analysis for \textit{LIANet}'s Sentinel-2 image generation.} In summary, our models perform decent (purple dash-dotted line) to excellent (red dashed line) with a distribution of PSNR-values derived from 100 randomly sampled spatial patches of size 128x128 pixels over the areas $\mathcal{A}_{0}$, $\mathcal{A}_{+}$, and $\mathcal{A}_{++}$. The label \textit{LIANet-Large}$_{\mathcal{A}_{++}}$ for instance, refers to the \textit{LIANet-Large} model that is pretrained on area  $\mathcal{A}_{++}$. Blue and orange labels correspond to PSNR statistics for all Sentinel-2 bands and RGB channels, respectively. Dots indicate minimum and maximum values, bars cover the $1.5\sigma$ interval, the cross marks the mean, and the open circle indicates the median. The bold solid blue and orange lines provide a logarithmic interpolation PSNR(bpp)$~\sim P_3\left[\log(1+\text{bpp})\right]$ of the median PSNR (RGB) values with $Q_\text{\it LIANet}=4$ where we fixed PSNR(bpp=0)=0. $P_3[z]$ denotes a polynomial of third degree in $z$. A vertical dashed line (gray) indicates the bpp of the raw Sentinel-2 data in JPEG2000 format as a reference.}
    \label{fig:lianetPSNR}
\end{figure}
The \textit{peak-signal-to-noise-ratio} with the definition of spatio-spectral averaging, $\langle\cdot\rangle$, reads
\begin{equation}
\text{PSNR}(\mathbf{I}\vert\mathbf{S}) = 10 \cdot \log_{10} \left( \frac{\max\mathbf{I}^2}{\langle(\mathbf{I}-\mathbf{S})^2\rangle} \right)
\quad.
\end{equation}
It quantifies the deviation of the \textit{LIANet}-generated image $\mathbf{I}$ against the corresponding Sentinel-2 L2A data $\mathbf{S}$ by computing the logarithm of the ratio: maximal image amplitude over the mean-squared error of reconstruction. In the limit where $\mathbf{I}\to\mathbf{S}$, PSNR$\to\infty$. If the \textit{LIANet} image strongly deviates from the Sentinel-2 reference, PSNR$\to-\infty$. When the root mean square error (RMSE) $\sqrt{\langle(\text{I}-\text{S})^2\rangle}\gtrsim\max\text{I}$ dominates over the maximum image signal strength $\vert\text{I}\vert$, PSNR is negative. At PSNR=0, the RMSE is equal to the maximal amplitude of pixel values in the reconstructed image. For a maximum signal ten times stronger than the error, PSNR$\approx$20. For excellent reconstruction where PSNR$\gtrsim$40, the error is at least 100 times weaker than the maximum signal.

\Cref{fig:lianetPSNR} provides PSNR as a function of bpp for various flavors of \textit{LIANet}.
Our results demonstrate that the reconstruction quality reaches acceptable (PSNR>30, signal approx.\ 30 times stronger than mean error) to excellent (PSNR>40) levels with minor inaccuracies, with the latter often imperceptible to the human eye.

\begin{figure*}[!]
    \centering
    \includegraphics[width=0.95\linewidth]{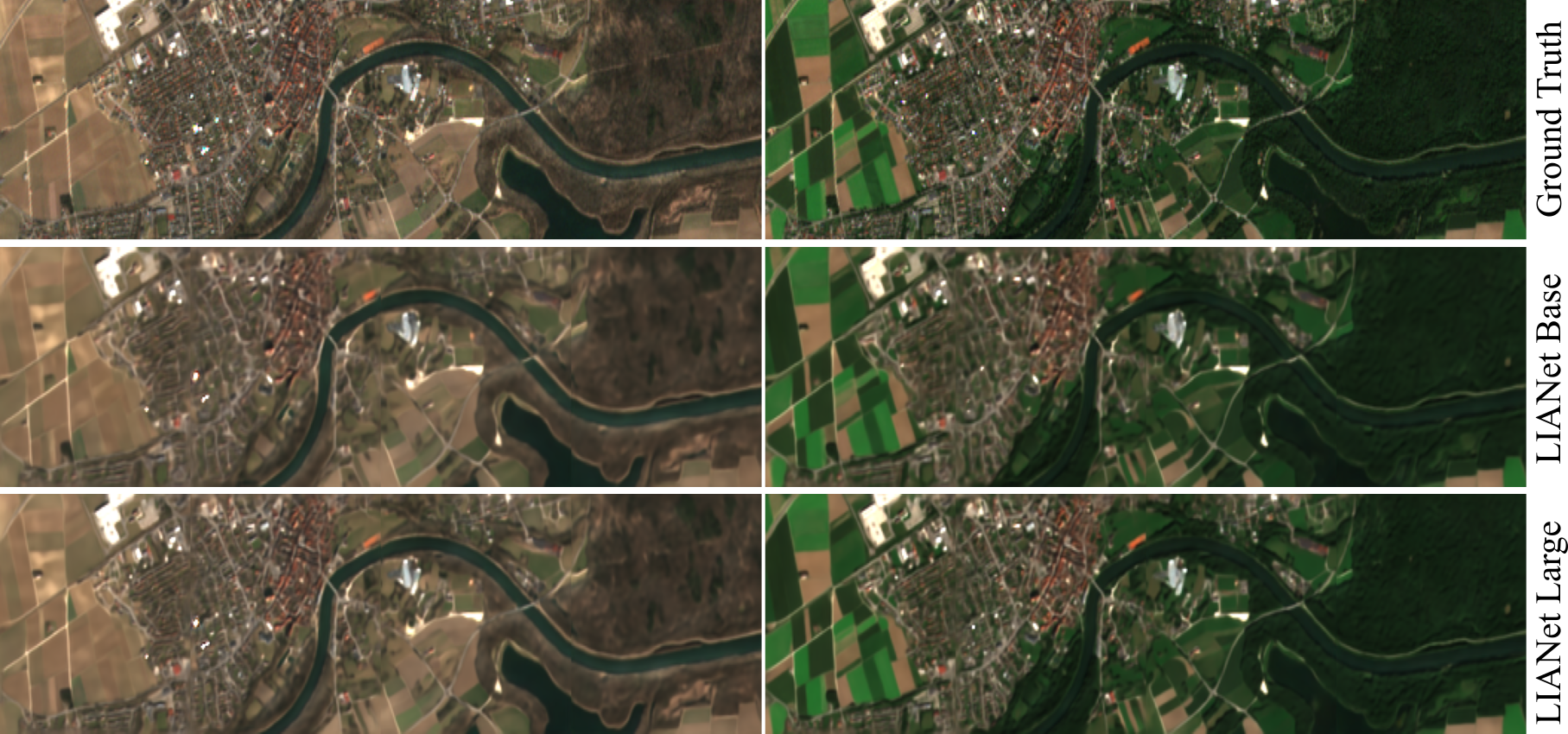}\\
    \vspace{0.1cm}
    \includegraphics[width=0.95\linewidth]{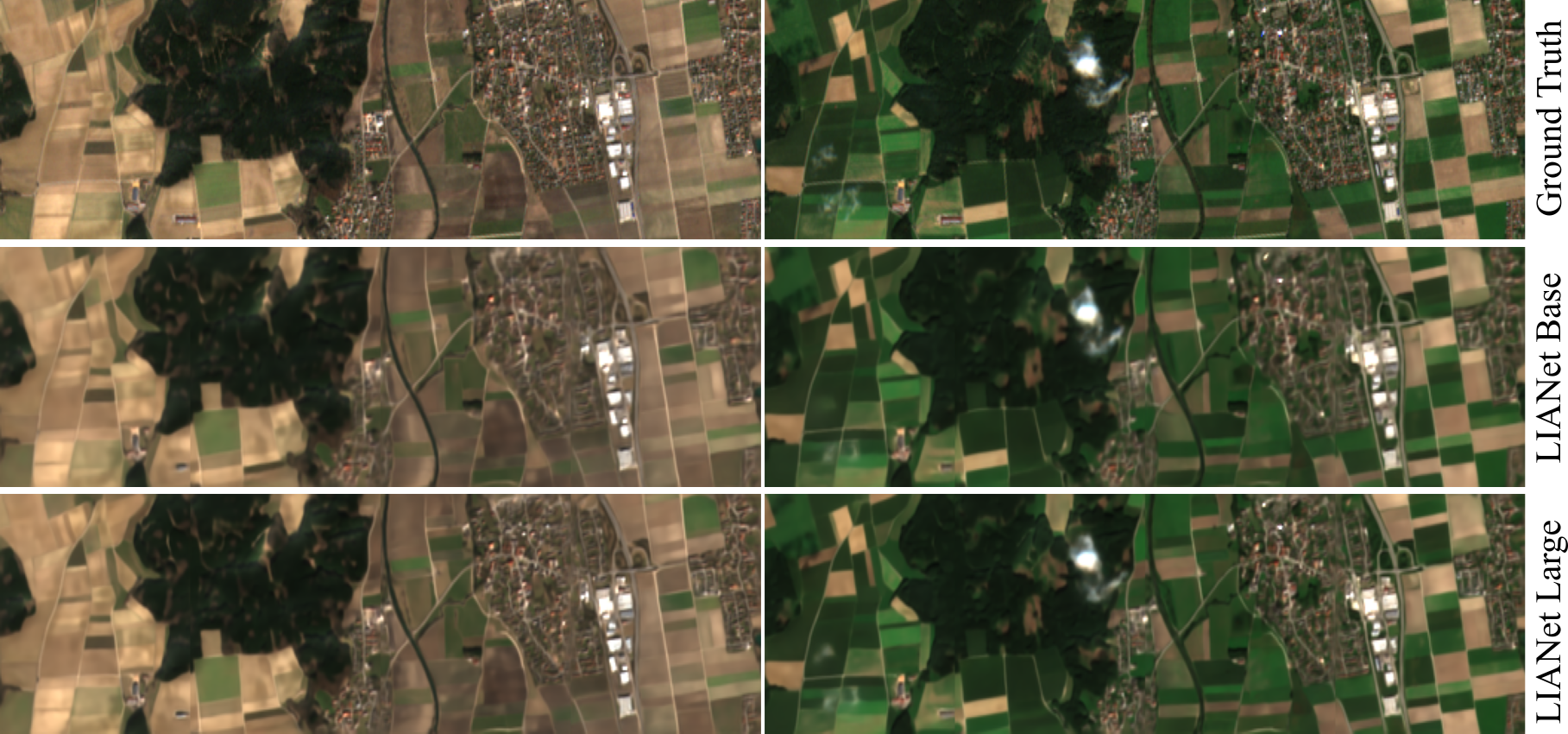}
    \caption{Two examples of reconstruction quality of \textit{LIANet-Base} and \textit{LIANet-Large} on $\mathcal{A}_{0}$ area. It can be seen that \textit{LIANet-Large} has an improved performance in reconstructing the high-frequency details, including buildings within the city. It can also be seen that cloudy areas are reconstructed as they are present in the Sentinel-2 image.}
    \label{fig:more_reconstruction_visual_samples}
\end{figure*}

The advancement in neural compression will guide future design updates of \textit{LIANet} to improve compression while maintaining downstream task utility. Based on the compression--quality trade-off summarized in \cref{fig:lianetPSNR}, the best model to pick is currently \textit{LIANet-Large}$_{\mathcal{A}_{+}}$: it provides enough parameter capacity to properly model a mid-size area of $\vert\mathcal{A}_+\vert=5000 km^2$ for (close to) excellent reconstruction across the entire Sentinel-2 spectral bands while keeping variability low\footnote{The standard deviation of all models is about $\Delta$PSNR$\gtrsim2$ around PSNR=40 which corresponds to signal-to-noise fluctuations of about two orders of magnitude.}, \ie, the model deals well with diverse remotely sensed scenes. To further assess the quality of reconstructed images by two \textit{LIANet} variants on $\mathcal{A}_{0}$, we illustrate two more examples in \cref{fig:more_reconstruction_visual_samples} with high-frequency details as well as clouds present in the scene. Each of the reconstructed images in \cref{fig:more_reconstruction_visual_samples} is generated with four forward passes and then stitching the patches horizontally.


\end{document}